\documentclass{article} 
\usepackage{iclr2027_conference,times}

\usepackage{amsmath,amsfonts,bm}

\def\eqref#1{equation~\ref{#1}}

\def\1{\bm{1}}

\DeclareMathAlphabet{\mathsfit}{\encodingdefault}{\sfdefault}{m}{sl}
\SetMathAlphabet{\mathsfit}{bold}{\encodingdefault}{\sfdefault}{bx}{n}

\usepackage{capt-of}
\usepackage{diagbox}

\definecolor{C0}{rgb}{0.121569, 0.466667, 0.705882}
\definecolor{C1}{rgb}{1.000000, 0.498039, 0.054902}
\definecolor{C2}{rgb}{0.172549, 0.627451, 0.172549}
\definecolor{C3}{rgb}{0.839216, 0.152941, 0.156863}
\definecolor{C4}{rgb}{0.580392, 0.403922, 0.741176}
\definecolor{C5}{rgb}{0.549020, 0.337255, 0.294118}
\definecolor{C6}{rgb}{0.890196, 0.466667, 0.760784}
\definecolor{C7}{rgb}{0.498039, 0.498039, 0.498039}
\definecolor{C8}{rgb}{0.737255, 0.741176, 0.133333}
\definecolor{C9}{rgb}{0.090196, 0.745098, 0.811765}
\definecolor{trolleygrey}{rgb}{0.5, 0.5, 0.5}
\definecolor{BrickRed}{rgb}{0.6,0,0}
\definecolor{RoyalBlue}{rgb}{0,0,0.8}
\definecolor{Tdgreen}{rgb}{0,0.4,0.7}
\definecolor{pinegreen}{rgb}{0.0, 0.47, 0.44}
\definecolor{cornellred}{rgb}{0.7, 0.11, 0.11}
\definecolor{cadmiumgreen}{rgb}{0.0, 0.42, 0.24}
\definecolor{spirodiscoball}{rgb}{0.06, 0.75, 0.99}
\definecolor{mylightblue}{rgb}{0.85, 0.90, 0.94}
\definecolor{maroon}{cmyk}{0,0.87,0.68,0.32}
\usepackage{pifont}
\definecolor{cfg}{rgb}{0.906, 0.435, 0.318}
\definecolor{cfgpp}{rgb}{0.165, 0.616, 0.561}
\definecolor{cfgnull}{rgb}{0.208, 0.565, 0.953}

\usepackage{hyperref}
\usepackage{url}
\usepackage{graphicx}
\usepackage{colortbl}
\usepackage{booktabs}
\usepackage[normalem]{ulem}

\usepackage{wrapfig}
\usepackage{tcolorbox}

\title{GeoFix: Geometry Fixing for 4D Consistent Generation}

\author{
\textbf{Yeobin Hong}\textnormal{\textsuperscript{1}} \hspace{1em}
\textbf{Hyojun Go}\textnormal{\textsuperscript{2}} \hspace{1em}
\textbf{Dohun Lee}\textnormal{\textsuperscript{1}} \hspace{1em}
\textbf{Jeongjae Lee}\textnormal{\textsuperscript{1}} \hspace{1em}
\textbf{Sunghwan Hong}\textnormal{\textsuperscript{2}} \\
\ \textbf{Federico Tombari}\textnormal{\textsuperscript{3}} \hspace{0.1em}
\textbf{Zhaochong An}\textnormal{\textsuperscript{4}} \hspace{0.1em}
\textbf{Jong Chul Ye}\textnormal{\textsuperscript{1,$\dagger$}} \hspace{0.1em}
\textbf{Konrad Schindler}\textnormal{\textsuperscript{2,$\dagger$}} \hspace{0.1em}
\textbf{Dominik Narnhofer}\textnormal{\textsuperscript{2,$\dagger$}} \\[3pt]
\ \textnormal{\textsuperscript{1}KAIST \hspace{1em} \textsuperscript{2}ETH Zürich \hspace{1em} \textsuperscript{3}Google \hspace{1em} \textsuperscript{4}EPFL} \\ 
\ \small $^{\dagger}$Corresponding authors
}

\iclrfinalcopy 

\begin{document}

\maketitle
\begingroup
\renewcommand{\thefootnote}{}
\footnotetext{Project Page: \url{https://yeobinhong.github.io/GeoFix/}}
\addtocounter{footnote}{-1}
\endgroup
\lhead{Preprint.}

\begin{figure}[h]
    \centering
    \vspace{-0.3cm}
    \includegraphics[width=\linewidth]{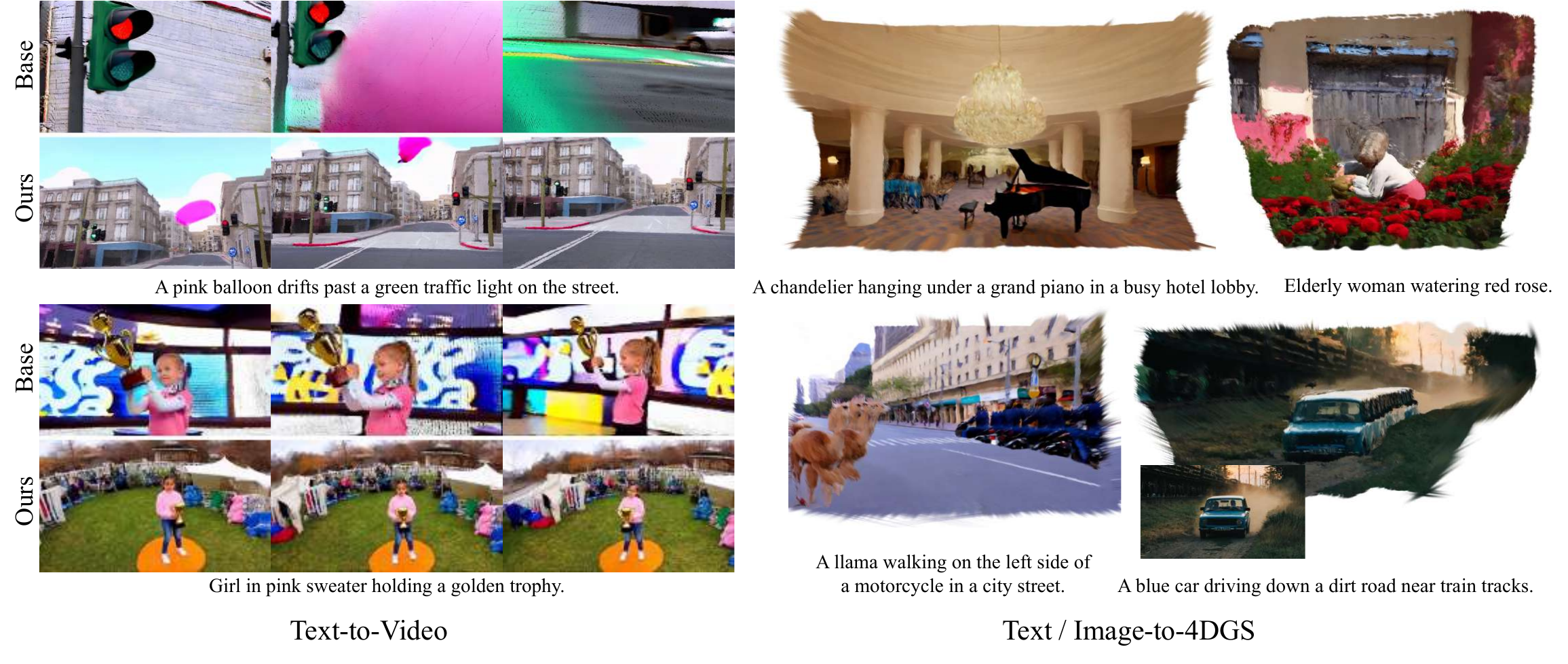}
    \vspace{-0.7cm}
    \caption{\textbf{Generated 4DGS and rendered videos from GeoFix}. Our geometric consistency reward yields more coherent 4D representations through both training and inference-time alignment.}
    \label{fig:represent}
\end{figure}

\begin{abstract}
Video diffusion models increasingly serve as priors for 4D generation, but the generated videos often lack geometric consistency.
Recent approaches address this through post-training by designing geometric rewards, yet this can degrade texture and fine detail,
requiring auxiliary appearance rewards and careful reward balancing.
To address this challenge, we propose GeoFix, a unified framework that improves geometric consistency and visual quality with a single geometric reward, applicable to both training and inference-time reward alignment.
Our reward measures cross-view depth consistency in static scene regions, targeting structural inconsistencies without directly constraining appearance or penalizing valid object motion. 
Crucially, depth agreement increases with camera field-of-view overlap even in real videos and 4D datasets, allowing a smaller camera movement to
be mistaken for better geometry. 
We account for this effect by calibrating the reward against the overlap-dependent trend.
GeoFix achieves a $39\%$ improvement in WorldScore 3D consistency and a $4\%$ increase in VBench Quality Score.
\end{abstract}
  
\section{Introduction}
Video diffusion models (VDMs)~\citep{wan2025wanopenadvancedlargescale,yang2024cogvideox} have achieved impressive visual fidelity by learning from large-scale video data.
However, their outputs can contain geometric inconsistencies, such as background structures deforming or drifting as the camera moves~\citep{duan2025worldscoreunifiedevaluationbenchmark}.
These artifacts limit their use as priors for 4D generation~\citep{wu2024cat4d,wang20254realvideov2fusedviewtimeattention,yang2026neoverseenhancing4dworld}, which requires scene structure and dynamics to remain coherent across time and viewpoints.
Improving this geometric coherence is essential for turning visually compelling videos into reliable dynamic scene representations.

Recent work has therefore explored reward-based post-training to improve the geometric consistency of pretrained video models.
One line of work provides indirect geometric feedback by assessing the visual quality or plausibility of rendered views using learned preference models or VLMs~\citep{go2026textto3dstitchingmultiviewreconstruction,wang2026worldr1reinforcing3dconstraints}.
Such feedback is informative, but it conflates geometric inconsistency with appearance and rendering fidelity, which limits its ability to target geometric errors.

Other rewards evaluate consistency more directly through RGB reconstruction errors~\citep{go2026textto3dstitchingmultiviewreconstruction,du2026videogpadistillinggeometrypriors}, epipolar constraints~\citep{kupyn2026epipolargeometryimprovesvideo}, depth agreement~\citep{an2026vggrpoworldconsistentvideogeneration}, or agreement between geometric reprojections and tracked correspondences~\citep{yin2026vigorvideogeometryorientedreward}.
However, optimizing these rewards can favor outputs that are easier to match or reconstruct, improving consistency at the expense of visual detail~\citep{go2026textto3dstitchingmultiviewreconstruction,kupyn2026epipolargeometryimprovesvideo}.
To mitigate these effects, existing methods introduce additional rewards that penalize nearly static videos~\citep{kupyn2026epipolargeometryimprovesvideo,yin2026vigorvideogeometryorientedreward} or reward visual quality~\citep{go2026textto3dstitchingmultiviewreconstruction}, or use explicit preference-data filtering~\citep{du2026videogpadistillinggeometrypriors}. 
This reliance on auxiliary rewards or preference-data filtering complicates alignment design, requiring additional choices about reward weights or sample-selection thresholds.

Therefore, current rewards that evaluate 3D consistency fall short of the broader goal of geometric alignment: correcting structural inconsistencies without harming the visual richness of pretrained models.
Since geometric inconsistencies are themselves visual defects, we argue that correcting them should also improve video quality. 
This motivates our central question: \textit{can a single geometric reward improve both geometric consistency and video quality without auxiliary rewards?}

In this paper, we introduce \textbf{GeoFix}, a framework that improves both geometric consistency and video quality using a single geometric reward. 
GeoFix combines three components: our \textbf{FO}V-\textbf{CAL}ibrated geometric reward (\textbf{FOCAL}), a stitched 4D reconstructor, and alignment methods.

At the core of GeoFix, FOCAL measures cross-view depth consistency to target scene structure without directly requiring colors or textures to match.
We restrict the comparison to static scene regions to avoid penalizing valid object motion as geometric inconsistency. 
A key observation behind FOCAL is that depth agreement is not determined by geometry alone. Even clean real videos score higher when the camera field-of-view (FOV) overlap is larger.
FOCAL therefore calibrates depth agreement against the score expected for a given overlap, using the relationship estimated from clean real videos and 4D datasets and kept fixed during optimization.
This reduces the extent to which greater overlap alone is rewarded as better geometry.

To compute FOCAL directly from generated video latents, we adapt model stitching~\citep{go2026textto3dstitchingmultiviewreconstruction} to a pretrained 4D reconstructor~\citep{yang2026neoverseenhancing4dworld}.
This bypasses RGB decoding and the reconstructor's early layers, and is more robust to the latent noise introduced during generation~\citep{go2026textto3dstitchingmultiviewreconstruction}.
The resulting model supplies depth and camera estimates for reward computation while producing an explicit 4D Gaussian representation from the latent.

FOCAL can be integrated into existing methods for both post-training and inference-time alignment.
We demonstrate this within GeoFix using DiffusionNFT~\citep{zheng2025diffusionnft} for post-training, and best-of-$N$ and DPS-style gradient-based guidance~\citep{chung2023diffusion} at inference time.
Experiments on WorldScore~\citep{duan2025worldscoreunifiedevaluationbenchmark} and VBench~\citep{huang2023vbenchcomprehensivebenchmarksuite} show that GeoFix outperforms baselines in geometric consistency and video quality, while also improving camera accuracy.
We further demonstrate its effectiveness in 4D Gaussian scene generation, producing more stable background geometry and coherent object motion.

\section{Related Work}

  
\paragraph{Diffusion Model Alignment.}
Reward alignment adapts pretrained models using rewards or preferences through post-training or inference-time steering. 
Post-training methods include differentiable reward optimization~\citep{clark2023directly}, reward-weighted regression~\citep{lee2023aligningtexttoimagemodelsusing}, policy-gradient methods~\citep{black2024ddpo,fan2023dpok,liu2025flowgrpotrainingflowmatching}, and direct preference optimization~\citep{wallace2023diffusionmodelalignmentusing}. 
Inference-time approaches include best-of-$N$ selection~\citep{Ma_2025_CVPR}, particle resampling~\citep{singhal2025generalframeworkinferencetimescaling}, gradient-free guidance~\citep{kim2025free2guidetrainingfreetexttovideoalignment}, and sampling-noise optimization~\citep{tang2024inferencetimealignmentdiffusionmodels}. Our work focuses on geometric reward design, whose effectiveness we evaluate through DiffusionNFT post-training~\citep{zheng2025diffusionnft}, best-of-$N$ selection and DPS-style gradient guidance~\citep{chung2023diffusion}.

\paragraph{Geometric Rewards for Video and 4D Generation.}
Building on diffusion alignment methods, recent works develop rewards to improve geometric consistency in video and 4D generation. 
These approaches draw on two broad types of feedback: indirect quality assessments and direct geometric consistency measurements. 
Indirect feedback assesses the perceived quality or plausibility of generated and rendered views using preference models or VLMs~\citep{go2026textto3dstitchingmultiviewreconstruction,wang2026worldr1reinforcing3dconstraints}. 
Such feedback provides useful perceptual supervision, but its dependence on appearance and rendering fidelity limits its ability to isolate structural errors. 

Direct feedback measures reconstruction errors~\citep{go2026textto3dstitchingmultiviewreconstruction,du2026videogpadistillinggeometrypriors,park2025steerx,wang2026worldr1reinforcing3dconstraints,ge2026campilot}, epipolar constraints~\citep{kupyn2026epipolargeometryimprovesvideo}, geometric correspondence consistency~\citep{yin2026vigorvideogeometryorientedreward,gu2025geco}, or depth agreement~\citep{an2026vggrpoworldconsistentvideogeneration}. 
However, directly optimizing consistency does not ensure better visual quality. 
For example, VIST3A~\citep{go2026textto3dstitchingmultiviewreconstruction} reports blurred outputs when reconstruction consistency is optimized without a quality reward. Epipolar-DPO~\citep{kupyn2026epipolargeometryimprovesvideo} reports oversaturation with a particular correspondence estimator.
Existing methods use auxiliary objectives~\citep{go2026textto3dstitchingmultiviewreconstruction,wang2026worldr1reinforcing3dconstraints,an2026vggrpoworldconsistentvideogeneration,kupyn2026epipolargeometryimprovesvideo,yin2026vigorvideogeometryorientedreward,li2026geo} or preference-data filtering~\citep{du2026videogpadistillinggeometrypriors}, complicating alignment design, requiring additional choices about reward weights or sample-selection thresholds. 

We address this challenge by designing a single geometric reward that improves both geometric consistency and video quality without auxiliary rewards, eliminating the need to balance multiple reward objectives.


\section{Method}

\begin{figure*}[t]
    \centering
    \vspace{-0.3cm}
    \includegraphics[width=\linewidth]{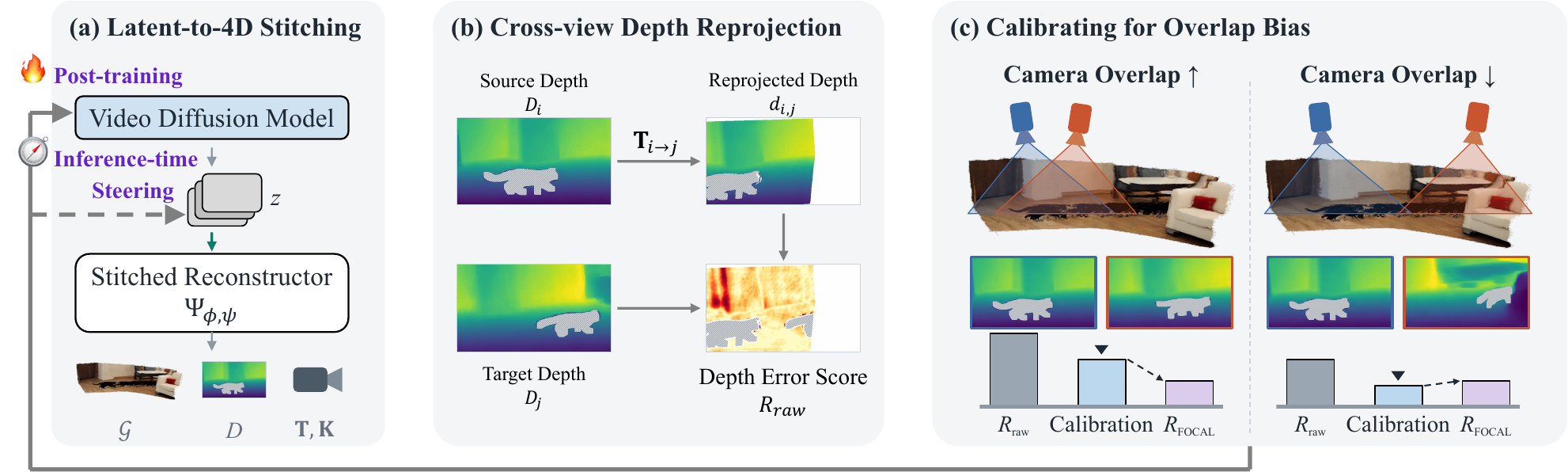}
    \vspace{-0.5cm}
    \caption{\textbf{Overview of GeoFix.} (a) FOCAL is computed directly from video latents using a stitched 4D reconstructor. (b) We measure cross-view depth reprojection error in static regions and (c) calibrate it to remove overlap bias.}
    \label{fig:method}
\end{figure*}

Our goal is to improve video quality by correcting geometric inconsistencies without unnecessarily constraining the rich textures and fine visual details of pretrained video models. 
To this end, we present \textbf{GeoFix}, a unified framework that aligns video diffusion models with a geometric reward computed directly on video latents (Figure~\ref{fig:method}). 
The proposed reward, \textbf{FOCAL} (\textbf{FO}V-\textbf{CAL}ibrated geometric reward), measures cross-view depth agreement in static scene regions. 
We compute FOCAL using geometry obtained from video latents through a stitched 4D reconstructor (Section~\ref{sec:4d_stitching}). 
We identify an overlap bias in raw depth-consistency scores (Section~\ref{sec:observation}). 
FOCAL removes this bias by calibrating the score against the depth agreement expected at a given overlap (Section~\ref{sec:3d_reward}). 
Finally, GeoFix applies to \textit{any} reward alignment method without modification, which we demonstrate across training and inference-time alignment (Section~\ref{sec:reward_alignment}).


\subsection{Latent-to-4D Model Stitching}
\label{sec:4d_stitching}

Our geometric reward requires reliable depth and camera estimates,
which must be computed repeatedly during video alignment.
We obtain these estimates directly from video latents by adapting
model stitching~\citep{go2026textto3dstitchingmultiviewreconstruction,an2026vggrpoworldconsistentvideogeneration}
to a pretrained 4D reconstructor~\citep{yang2026neoverseenhancing4dworld}.
This bypasses the VAE decoder and the initial reconstruction layers,
avoiding their computational cost.
The choice is further motivated by VIST3A's finding that stitched
3D reconstruction is more robust to latent perturbations than
sequential RGB decoding and reconstruction~\citep{go2026textto3dstitchingmultiviewreconstruction}.
Beyond providing geometric feedback, the stitched model produces an explicit 4D Gaussian representation from the same video latent.

Let $\mathcal{F}$ denote the pretrained 4D reconstructor with $L$ layers.
A lightweight stitching layer $\mathcal{S}_{\phi}$ maps a video latent
$z$ into the feature space at the output of layer $k$.
We initialize a trainable reconstruction module $\mathcal{R}_{\psi}$
from the remaining layers $\mathcal{F}_{k+1:L}$.
The stitched reconstructor is then defined as
\begin{equation}
    \Psi_{\phi,\psi}(z)
    =
    \mathcal{R}_{\psi}\bigl(\mathcal{S}_{\phi}(z)\bigr)
    =
    \left(
        \mathcal{G},
        \left\{
            D_f,\mathbf{T}_f,\mathbf{K}_f
        \right\}_{f=1}^{F}
    \right),
    \label{eq:latent_to_4d}
\end{equation}
where $\mathcal{G}$ denotes the reconstructed 4D Gaussian representation,
$F$ is the number of frames, $D_f$, $\mathbf{T}_f$, and
$\mathbf{K}_f$ denote the depth map, camera-to-world pose, and camera
intrinsics of frame $f$, respectively.

We train $\phi$ and $\psi$ jointly by output distillation. For a video $x$, the frozen teacher $\mathcal{F}$ provides the target $\mathcal{F}(x)$. 
The stitched reconstructor predicts from the latent $\mathcal{E}(x)$, where $\mathcal{E}$ is the frozen VAE encoder.
We minimize
\begin{equation}
    \mathcal{L}_{\mathrm{stitch}}(\phi,\psi)
    =
    \mathbb{E}_{x\sim\mathcal{D}}
    \left[
        \sum_{q\in\mathcal{Q}}
        \lambda_q
        \left\|
            \Psi^{q}_{\phi,\psi}(\mathcal{E}(x))
            -
            \mathcal{F}^{q}(x)
        \right\|_1
    \right],
    \label{eq:stitch_training}
\end{equation}
where $\mathcal{D}$ contains multi-view image sets and monocular videos of dynamic scenes.
$\Psi^{q}$ and $\mathcal{F}^{q}$ denote the $q$-th output tensor
(depth, poses, intrinsics, or Gaussian attributes) and $\lambda_q$ weights its distillation loss.
The resulting stitched reconstructor achieves higher reconstruction fidelity than the original RGB-input reconstructor (see Table~\ref{tab:novel_time_view}).


\subsection{Cross-View Depth Consistency Reward}
\label{sec:observation}

We base our geometric reward on cross-view depth consistency to constrain scene structure without directly restricting the model's freedom to generate rich textures and fine visual details.
View-dependent effects such as specular highlights make RGB an unreliable signal of geometric consistency~\citep{asim2025met3r}, and optimizing it can favor blurred outputs that are easier to match across views~\citep{go2026textto3dstitchingmultiviewreconstruction}.
Instead, depth reprojection evaluates cross-view agreement of the 3D structure, without requiring colors or textures to match or relying on point-tracking correspondences.

For each frame $i$, we unproject the depth $D_i(\mathbf{p})$ with $\mathbf{K}_i$ to obtain the center of the pixel-aligned Gaussians $\mathbf{X}_i(\mathbf{p})$ in camera $i$'s coordinates. 
For a frame pair $(i,j)$, we transform these points into the target camera using the relative pose $\mathbf{T}_{i\to j}=\mathbf{T}_j^{-1}\mathbf{T}_i$ and project them as
\begin{equation}
    \bigl(\mathbf{y}_{i,j}(\mathbf{p}),d_{i,j}(\mathbf{p})\bigr)
    =
    \Pi\bigl(
        \mathbf{T}_{i\to j}\cdot\mathbf{X}_i(\mathbf{p}),
        \mathbf{K}_j
    \bigr),
    \label{eq:warp}
\end{equation}
where $\Pi$ is the forward projection function, $\mathbf{y}_{i,j}(\mathbf{p})$ is the projected pixel location in frame $j$ and $d_{i,j}(\mathbf{p})$ is the target-camera depth.
Depth agreement is measured by comparing $d_{i,j}(\mathbf{p})$ with the target depth map $D_j$, bilinearly sampled at $\mathbf{y}_{i,j}(\mathbf{p})$.

Reprojection using camera transformations alone does not account for independent object motion.
To avoid treating valid motion as geometric inconsistency, we introduce a dynamic-region mask and restrict the comparison to static content in both views.
Specifically, we apply a pretrained motion segmentation network~\citep{huang2025segmentmotionvideos} to videos rendered from $\mathcal{G}$ to obtain per-frame dynamic masks.
We aggregate the remaining depth discrepancies as
\begin{equation}
    \mathrm{MSE}
    =
    \frac{1}{|\mathcal{C}|}
    \sum_{(i,j)\in\mathcal{C}}
    \frac{1}{|\mathcal{V}_{i,j}|}
    \sum_{\mathbf{p}\in\mathcal{V}_{i,j}}
    \left(
        D_j\bigl(\mathbf{y}_{i,j}(\mathbf{p})\bigr)
        -
        d_{i,j}(\mathbf{p})
    \right)^2,
    \label{eq:reproj_err}
\end{equation}
where $\mathcal{C}$ contains frame pairs with stride $s$.
The set $\mathcal{V}_{i,j}$ contains source pixels with valid depths whose projections lie within the target image, have positive target-camera depth, and fall in static regions in both views.
To obtain a reward that increases with depth agreement, we convert this error into a PSNR score:
\begin{equation}
    R_{\mathrm{raw}}
    =
    10\log_{10}
    \left(
        \frac{D_{\max}^{2}}{\mathrm{MSE}+\epsilon}
    \right),
    \label{eq:depth_psnr}
\end{equation}
where $D_{\max}$ is the maximum depth in the reconstructed clip and $\epsilon>0$ ensures numerical stability.

\begin{figure*}[t]
    \centering
    \vspace{-0.3cm}
    \includegraphics[width=0.8\linewidth]{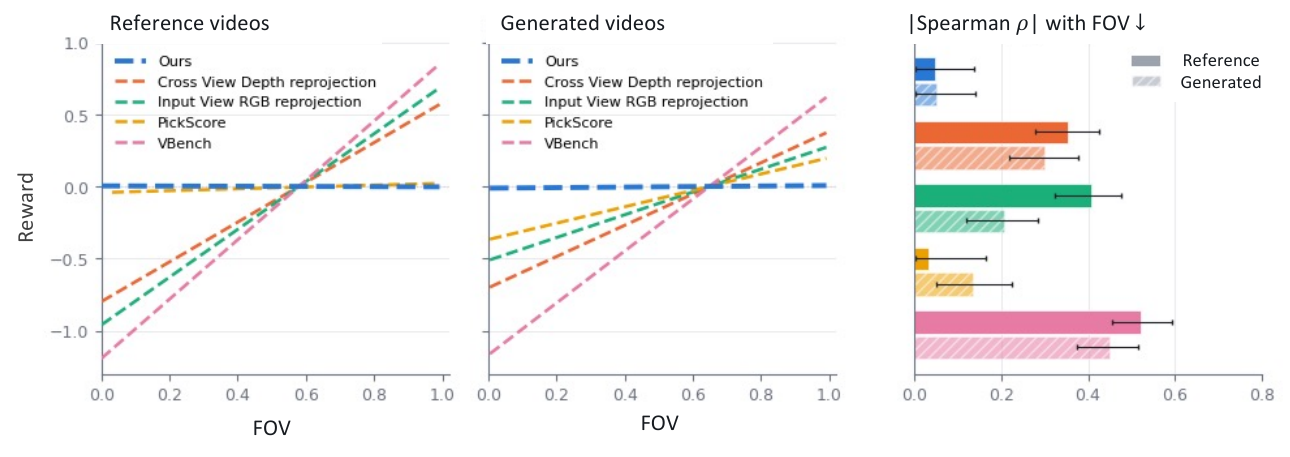}
    \vspace{-0.5cm}
    \caption{\textbf{Camera Correlation Analysis.} Existing direct and indirect geometric rewards vary with camera movement, creating a bias toward smaller viewpoint changes. Our FOV-calibrated reward remains invariant, providing a more reliable measure of geometric consistency.}
    \label{fig:camera_corr}
\end{figure*}

\paragraph{Observation: Overlap Bias.}
Although $R_{\mathrm{raw}}$ evaluates structural agreement, a higher score does not necessarily indicate better geometry. 
We find that even real videos and multi-view 4D datasets, which we use as reference videos, exhibit lower depth reprojection errors when camera overlap is larger (Figure~\ref{fig:camera_corr}). 
The same dependence holds for existing geometric rewards (Section~\ref{sec:fov}). 
We refer to this effect as overlap bias. These scores cannot disentangle geometric quality from camera overlap, so maximizing them favors larger camera overlap without improvement in scene geometry.

To quantify camera overlap, we use the area-weighted IoU of the two cameras' fields of view~\citep{lee2026memoryv2vmemoryaugmentedvideotovideodiffusion}. 
We sample a reference sphere centered at the first camera on a uniform azimuth--elevation grid, and let $\mathcal{A}_f$ denote the sampled points within camera $f$'s field of view.
The overlap between the first and last cameras is
\begin{equation}
    \Omega
    =
    \frac{
        ||\mathcal{A}_1\cap\mathcal{A}_F||
    }{
        ||\mathcal{A}_1\cup\mathcal{A}_F||
    },
    \label{eq:fov_overlap}
\end{equation}
where $||\cdot||$ denotes a count weighted by the cosine of elevation to account for spherical area.

\subsection{FOCAL: FOV-Calibrated geometric reward}
\label{sec:3d_reward}

Motivated by this observation, we use reference videos to estimate the raw depth consistency score expected at a given FOV overlap. We then evaluate each generated video relative to this overlap-dependent reference.
We estimate the reference by fitting the relationship between the raw depth score and FOV overlap:
\begin{equation}
    R_{\mathrm{raw}}
    =
    \alpha
    +
    \beta\log_{10}\max(\Omega,\epsilon_{\mathrm{fov}})
    +
    \xi,
    \label{eq:raw_fit}
\end{equation}
where $\alpha$ is the intercept, $\beta$ is the slope,
and $\xi$ is the residual.
The small constant $\epsilon_{\mathrm{fov}}>0$ ensures numerical
stability at zero overlap.
To compare a generated video's score with this expected score,
we subtract the fitted overlap-dependent term.
The fitted intercept is constant and does not affect reward
maximization, so we omit it from the correction:
\begin{equation}
    R_{\mathrm{FOCAL}}
    =
    R_{\mathrm{raw}}
    -
    \hat{\beta}\log_{10}\max(\Omega,\epsilon_{\mathrm{fov}}),
\end{equation}
where $\hat{\beta}$ is the slope estimated from reference videos.
A high reward therefore indicates strong depth agreement relative
to the expected score for the video's camera overlap.
Equivalently, defining the calibration coefficient $\gamma=\hat{\beta}/10$, we obtain FOCAL (\textbf{FO}V-\textbf{CAL}ibrated geometric reward):
\begin{equation}
    R_{\mathrm{FOCAL}}
    =
    10\log_{10}
    \left(
        \frac{D_{\max}^{2}}
        {(\mathrm{MSE}+\epsilon)
        \max(\Omega,\epsilon_{\mathrm{fov}})^{\gamma}}
    \right).
    \label{eq:3d_reward}    
\end{equation}

We estimate the calibration coefficient $\gamma$ only from reference videos and keep it fixed throughout alignment.
On held-out real and generated videos, the corrected score--overlap trend has a near-zero regression slope, showing that the correction also reduces overlap bias beyond the calibration set. FOCAL thus accounts for viewing conditions within the geometric reward itself, without introducing auxiliary rewards.

\subsection{GeoFix: Reward-Based Video Alignment}
\label{sec:reward_alignment}

We refer to the resulting framework as GeoFix, a unified reward alignment scheme that uses FOCAL to fine-tune the VDM and to select or guide samples at inference time.
For a video latent $z$, we write $R_{\mathrm{FOCAL}}(z)$ for the reward of Eq.~\ref{eq:3d_reward} computed from the geometry $\Psi_{\phi,\psi}(z)$ predicted by the stitched reconstructor. 
$R_{\mathrm{FOCAL}}(z)$ operates in the latent space, requires no VAE decoding and is differentiable in $z$. GeoFix is therefore agnostic to the choice of alignment method, including methods that require reward gradients. To demonstrate this, we instantiate GeoFix with representative methods: reward fine-tuning (DiffusionNFT~\citep{zheng2025diffusionnft}), zeroth-order sample selection (Best-of-$N$~\citep{beirami2025theoreticalguaranteesbestofnalignment}), and gradient guidance (DPS~\citep{chung2023diffusion}). $\Psi$ and $\gamma$ stay fixed and no auxiliary reward is used in any setting.

\paragraph{Training-time Alignment with DiffusionNFT.}
We fine-tune the VDM using DiffusionNFT.
We generate latents, score them with $R_{\mathrm{FOCAL}}(z)$, and use these scores as feedback for its forward-process flow-matching objective.

\paragraph{Inference-time Alignment.}
We keep the video model fixed and use the reward in two ways.

\textbf{1) Best-of-$N$.}
We generate $N$ complete video samples for each prompt, score their latents with $R_{\mathrm{FOCAL}}(z)$, and return the sample with the highest reward.

\textbf{2) DPS-style Guidance.}
We use the reward gradient to guide sampling following the gradient-based approach of DPS.
At each guided step, we evaluate the reward on the clean latent prediction $\hat{z}_{0\mid t}$ from $z_t$ and adjust the next latent in a direction that locally increases the reward:
\begin{equation}
    z_{t-1}
    =
    \tilde{z}_{t-1}
    +
    \eta_t\nabla_{z_t}R_{\mathrm{FOCAL}}\bigl(\hat{z}_{0\mid t}\bigr),
\end{equation}
where $\tilde{z}_{t-1}$ is the next latent from the unguided sampling and $\eta_t\geq0$ controls the guidance strength.

\section{Experiments}

\subsection{Experimental Setup}
\label{sec:exp_setup}
For training-time alignment, we use Wan2.2-TI2V-5B~\citep{wan2025wanopenadvancedlargescale} and jointly optimize a supervised loss on MultiCamVideo~\citep{bai2025recammastercameracontrolledgenerativerendering} and UltraVideo~\citep{xue2025ultravideo}, and a DiffusionNFT~\citep{zheng2025diffusionnft} loss using 64 rollouts per step generated from VideoUFO~\citep{wang2025videoufomillionscaleuserfocuseddataset} prompts and scored by $R_\mathrm{FOCAL}$. 
For inference-time alignment, we use Wan2.2-Fun-5B-Control-Camera.
Since both models share the same VAE, we use a fixed stitched reconstructor fine-tuned from NeoVerse~\citep{yang2026neoverseenhancing4dworld}.
Implementation details are provided in Appendix~\ref{sec:supp_impl}.

\subsection{VDM Training-time Alignment}
\label{sec:exp_train}

\begin{figure*}[t]
    \centering
    \vspace{-0.3cm}
    \includegraphics[width=\linewidth]{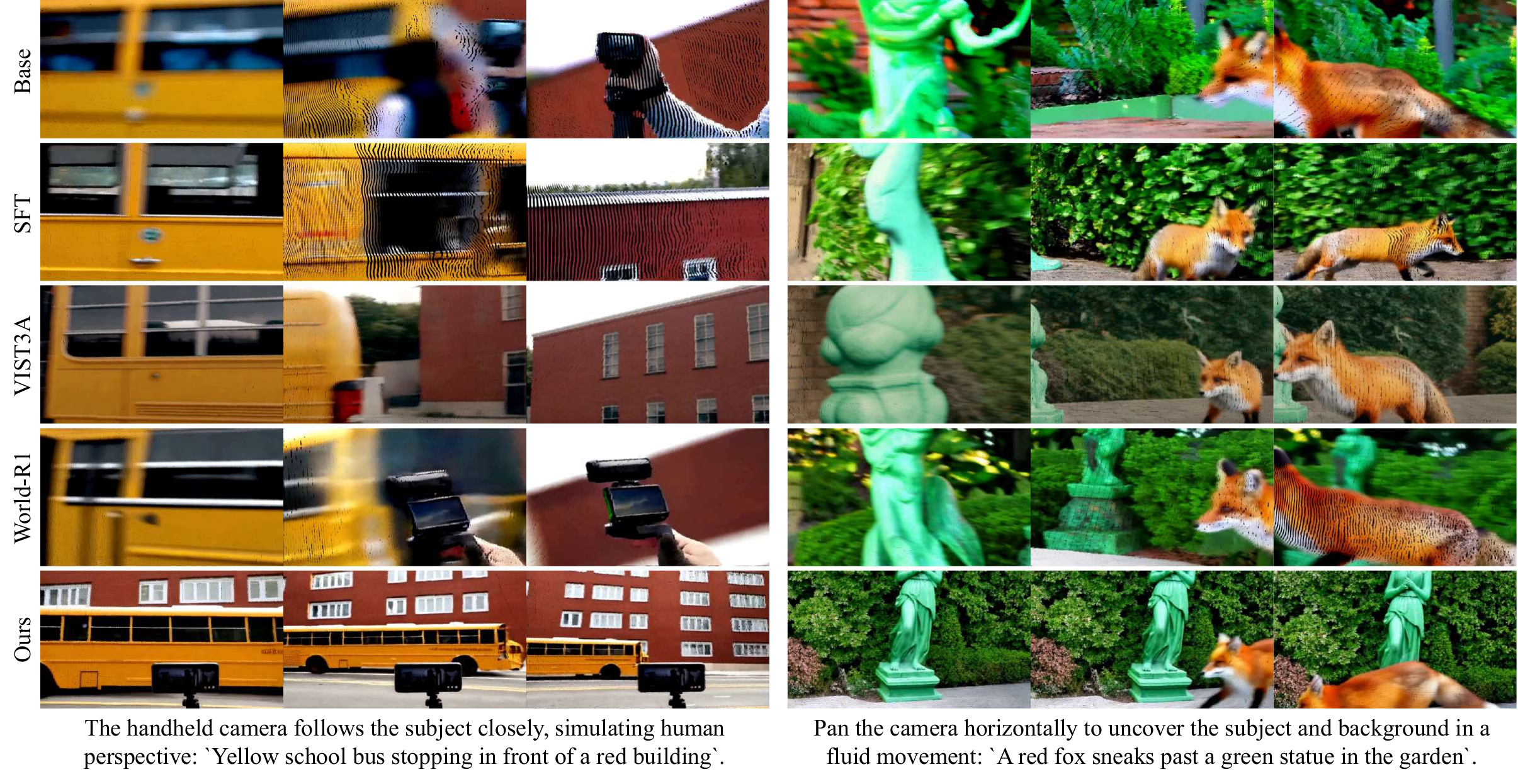}
    \vspace{-0.5cm}
    \caption{Qualitative comparison of 4DGS-rendered Text-to-Video generation.}
    \label{fig:qualitative_t2v}
\end{figure*}

\paragraph{Baselines, Evaluation Setting.}
We compare our method against all baselines on Wan2.2-TI2V-5B: base model, supervised fine-tuning (SFT), VIST3A~\citep{go2026textto3dstitchingmultiviewreconstruction} and World-R1~\citep{wang2026worldr1reinforcing3dconstraints}. All fine-tuned models use the same SFT dataset and training budget, and leverage the same stitched reconstructor. We generate a 4DGS scene for 200 T2V-CompBench~\citep{sun2025t2vcompbenchcomprehensivebenchmarkcompositional} prompts and evaluate on the rendered videos. We report on FOCAL for geometric consistency and WorldScore~\citep{duan2025worldscoreunifiedevaluationbenchmark} for world generation quality. 
We also compare against VGGRPO~\citep{an2026vggrpoworldconsistentvideogeneration}, which leverages a direct geometric reward. Since VGGRPO has released neither code nor checkpoints, we compare with the results reported in its paper. Following its protocol, we score the VDM decoded outputs for the standard VBench prompts. We report the relative improvement each method achieves over its own base model's results.

\begin{table}[t]
    \centering
    \vspace{-0.5cm}
    \caption{Quantitative comparison of training-time reward alignment methods.}    
    \label{tab:t2v_quantitative_comparison}
    \resizebox{\textwidth}{!}{
    \begin{tabular}{lccccccccc}
        \hline
        & $\boldsymbol{R_{\mathrm{FOCAL}}}$ & \multicolumn{8}{c}{\textbf{WorldScore}} \\
        & & \multicolumn{2}{c}{Scene Consistency} & \multicolumn{3}{c}{Dynamics} & \multicolumn{2}{c}{Visual Quality} & Text Alignment \\
        \cmidrule(r){2-2} \cmidrule(r){3-4} \cmidrule(r){5-7} \cmidrule(r){8-9} \cmidrule(r){10-10}
        Method & Avg & 3D Cons. & Photo. Cons. & Motion Acc. & Motion Mag. & Motion Smooth. & Style Cons. & Sub. Qual. & Content Align. \\
        \cmidrule(r){1-1} \cmidrule(r){2-10}
        Base & 29.99 & 44.39 & 33.55 & 50.30 & 45.17 & 18.08 & 84.70 & 6.80 & 89.36 \\
        SFT & 31.02 & 50.11 & 38.47 & 52.76 & \textbf{48.93} & 19.90 & 89.00 & 9.80 & 87.17 \\
        VIST3A & \underline{31.06} & \underline{58.32} & \underline{47.00} & \underline{52.84} & 41.94 & \underline{25.62} & \textbf{96.63} & \underline{15.16} & \textbf{95.47} \\
        World-R1 & 30.49 & 42.26 & 31.64 & 51.35 & \underline{45.90} & 19.72 & 88.07 & 9.57 & 91.03 \\
        \rowcolor{blue!10} Ours & \textbf{33.01} & \textbf{61.60} & \textbf{51.06} & \textbf{55.94} & 35.44 & \textbf{26.08} & \underline{93.30} & \textbf{15.94} & \underline{91.40} \\
        \hline
    \end{tabular}
    }
\end{table}

\begin{table}[t]
    \centering
    \vspace{-0.5cm}
    \caption{Quantitative comparison of training-time reward alignment against VGGRPO. Parentheses show relative improvements over the base model.}
    \label{tab:video_quality_metrics}
    \resizebox{\textwidth}{!}{
    \begin{tabular}{lccccc}
        \hline
        & \multicolumn{5}{c}{\textbf{VBench}} \\
        Method & Subject Consistency & Background Consistency & Aesthetic Quality & Imaging Quality & Motion Smoothness \\
        \cmidrule(r){1-1} \cmidrule(r){2-6}
        Base & 0.9542 & 0.9528 & 0.5966 & 0.6733 & 0.9841 \\
        \rowcolor{gray!10} VGGRPO & 0.9644 (+1.07\%) & 0.9583 (+0.58\%) & 0.5991 (+0.42\%) & 0.6861 (+1.90\%) & 0.9895 (+0.55\%) \\        \hline        
        Base & 0.8741 & 0.9389 & 0.4679 & 0.5523 & 0.9589 \\ 
        \rowcolor{blue!10} Ours & \textbf{0.9317 (+6.59\%)} & \textbf{0.9601 (+2.25\%)} & \textbf{0.5272 (+12.67\%)} & \textbf{0.5979 (+8.25\%)} & \textbf{0.9802 (+2.22\%)} \\
        \hline
    \end{tabular}
    }
\end{table}

\begin{table}[t]
    \centering
    \vspace{-0.5cm}
    \caption{Quantitative comparison of inference-time reward alignment.}
    \label{tab:inference_time_steering}
    \resizebox{\textwidth}{!}{
    \begin{tabular}{lccccccccccc}
        \hline
        & \multicolumn{3}{c}{\textbf{Camera Accuracy}} & \multicolumn{8}{c}{\textbf{WorldScore}} \\
        & \multicolumn{3}{c}{} & \multicolumn{2}{c}{Scene Consistency} & \multicolumn{3}{c}{Dynamics} & \multicolumn{2}{c}{Visual Quality} & Text Alignment \\
        \cmidrule(r){2-4} \cmidrule(r){5-6} \cmidrule(r){7-9} \cmidrule(r){10-11} \cmidrule(r){12-12}
        Method & Rot. Err. $\downarrow$ & Trans. Err. $\downarrow$ & ATE Norm. $\downarrow$ & 3D Cons. & Photo. Cons. & Motion Acc. & Motion Mag. & Motion Smooth. & Style Cons. & Sub. Qual. & Content Align. \\
        \cmidrule(r){1-1} \cmidrule(r){2-12}
        None & 2.040 & 34.97 & 0.2563 & 75.43 & 76.06 & \textbf{44.81} & \textbf{31.93} & 23.65 & 95.47 & 20.87 & 94.10 \\
        \rowcolor{blue!10}
        Best-of-$N$ & \textbf{1.751} & \textbf{34.57} & \textbf{0.2507} & \textbf{78.00} & \textbf{78.33} & 44.59 & 31.15 & \textbf{24.61} & \textbf{96.08} & \textbf{21.86} & \textbf{94.66} \\
        \hline
        None & 2.040 & 34.97 & 0.2563 & 75.43 & 76.06 & \textbf{44.81} & \textbf{31.93} & 23.65 & 95.47 & 20.87 & 94.10 \\
        \rowcolor{blue!10} DPS & \textbf{1.977} & \textbf{34.63} & \textbf{0.2562} & \textbf{76.49} & \textbf{76.59} & 44.42 & 31.83 & \textbf{23.84} & \textbf{95.56} & \textbf{21.15} & \textbf{94.55} \\
        \hline
    \end{tabular}
    \vspace{-0.5cm}
    }
\end{table}

\paragraph{Results.}
GeoFix achieves the strongest geometric consistency among all baselines while also improving general video quality, without an auxiliary reward. As shown in Table~\ref{tab:t2v_quantitative_comparison}, it obtains the highest 3D consistency and photometric consistency on WorldScore. SFT alone yields only moderate gains, indicating that most of the improvement comes from the reward rather than from the supervised data. GeoFix also attains the best motion accuracy and motion smoothness, and remains superior or competitive on visual quality and text alignment. This supports our claim that correcting geometric inconsistencies, which are themselves visual defects, also improves video quality.
Furthermore, our method improves every VBench dimension by a larger relative margin than VGGRPO (Table~\ref{tab:video_quality_metrics}), with the largest gaps in visual quality. Note that none of these video quality metrics is used as a reward during training.

Qualitatively, our method produces the cleanest 4D scene renderings, with fewer artifacts and substantially reduced background drift and flickering compared with the baselines.
Figure~\ref{fig:qualitative_t2v} compares the videos rendered from the generated 4DGS scenes. Evaluating these renderings exposes geometric inconsistencies that can manifest as reconstruction and rendering artifacts. For the baselines, the reconstructor receives geometrically inconsistent videos, resulting in scenes with rendering artifacts.

\subsection{Inference-time Alignment}
\label{sec:exp_inference}

\paragraph{Baselines, Evaluation Setting.}
We plug our reward into two inference-time steering methods. We use $50$ NFEs, $N{=}20$ for Best-of-$N$ and guiding step $3$ to $17$ for DPS. 
We pair each of the 355 VBench-I2V~\citep{huang2024vbenchcomprehensiveversatilebenchmark} images with one of ten target camera trajectories. We measure camera accuracy between the camera trajectory recovered by the reconstructor and the target trajectory. We evaluate WorldScore for world generation quality on the rendered videos.

\paragraph{Results.}
Our proposed reward improves both camera accuracy and 4D scene quality for both first and zeroth order inference-time steering methods (Table~\ref{tab:inference_time_steering}). This indicates that our reward even provides a useful gradient signal through the stitched model. As in training-time alignment results, GeoFix not only improves consistency metrics but also visual quality and text alignment metrics. 
These improvements are accompanied by only a marginal reduction in motion accuracy and motion magnitude, whereas every camera accuracy and every other WorldScore aspect improve at the same time.

\subsection{4D Scene Generation}
\label{sec:exp_4d}

\paragraph{Baselines, Evaluation Setting.}

We compare image-to-4D generation with MoGe4D~\citep{zhang2026geometryawaresingleimage4dsynthesis} and 4DNeX~\citep{chen20254dnexfeedforward4dgenerative} on 100 images from VBench-I2V~\citep{huang2024vbenchcomprehensiveversatilebenchmark}. 
We render every generated 4D scene along an identical novel-view camera trajectory from ReCamMaster~\citep{bai2025recammastercameracontrolledgenerativerendering}. 
Following the baseline protocol, we fill disoccluded regions using MoGe4D’s 4D-ViSM identically across all methods. 
We then score the resulting novel-view videos with six complementary metrics. 
From VBench, we use subject consistency and temporal flickering for consistency and temporal stability of the rendered scene, dynamic degree for the amount of motion, aesthetic quality for visual quality. We use CLIP score~\citep{radford2021learning} for text alignment, and physical commonsense from VideoPhy-2~\citep{bansal2025videophy2challengingactioncentricphysical} for physical plausibility.

\paragraph{Results.}

\begin{figure*}[t]
    \centering
    \vspace{-0.3cm}
    \includegraphics[width=\linewidth]{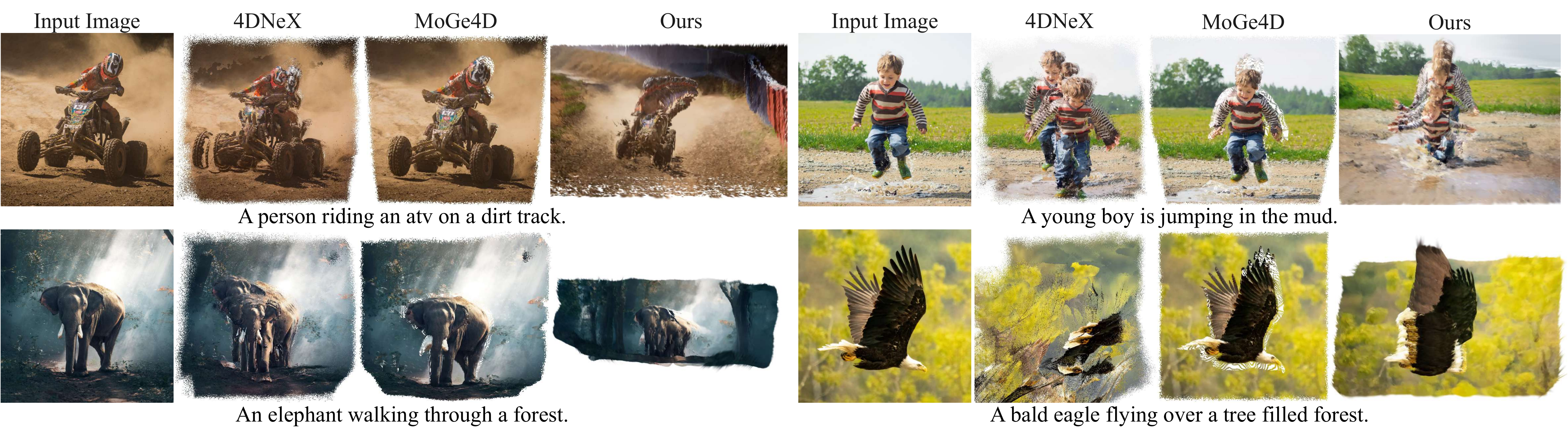}
    \vspace{-0.8cm}
    \caption{Qualitative comparison on Image-to-4D generation. We overlay 3D representations from multiple timesteps in a single view to visualize dynamic motion over time.}
    \label{fig:qualitative_i24d}
\end{figure*}

\begin{wrapfigure}{r}{0.3\textwidth}
  \begin{center}
\vspace{-1.0cm}
  \includegraphics[width=0.3\textwidth]{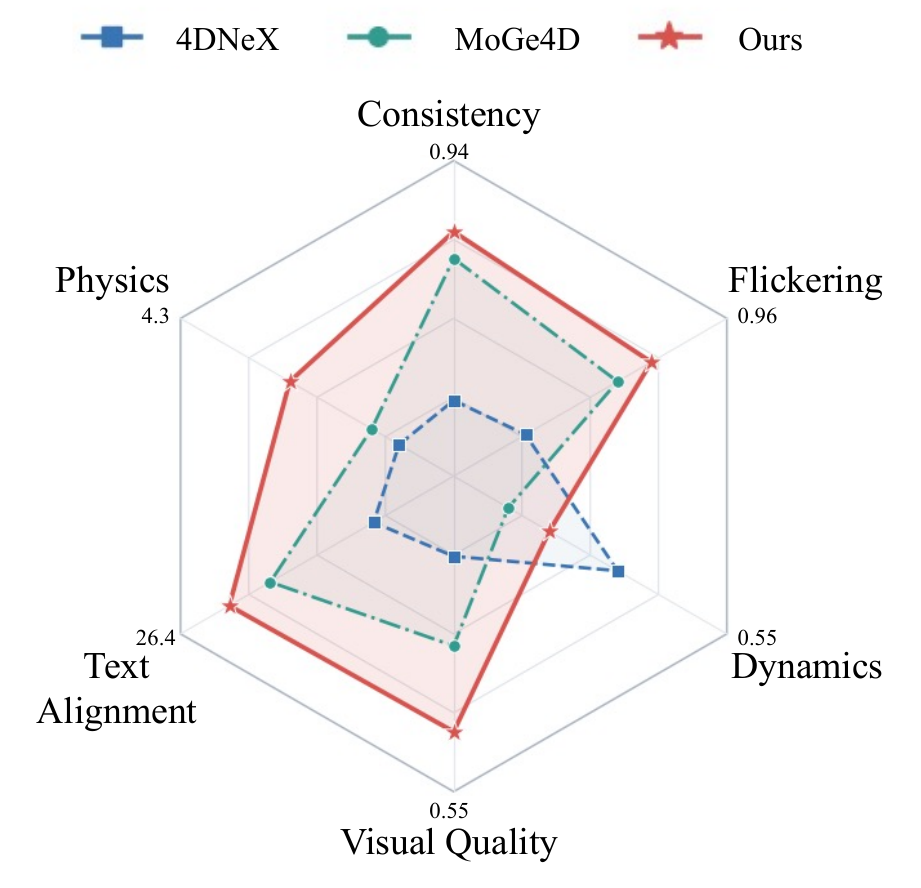}
  \end{center}
 \vspace{-0.6cm}
  \caption{Quantitative comparison on 4D Scene Generation. Higher is better; axes use different scales.}
  \vspace{-0.4cm}
  \label{fig:quantitative_i24d}
\end{wrapfigure}

GeoFix performs best overall and achieves the highest consistency scores (Figure~\ref{fig:quantitative_i24d}), indicating stable geometry under novel camera trajectories. 4DNeX attains the highest dynamic degree, but this largely arises from frame-to-frame structural inconsistency rather than valid object motion (Figure~\ref{fig:qualitative_i24d}), consistent with its poor consistency scores. MoGe4D exhibits the lowest dynamic degree, indicating insufficient motion for 4D synthesis. Ours better preserves scene structure while maintaining meaningful dynamics.

Figure~\ref{fig:qualitative_i24d} shows that GeoFix produces coherent 4D scenes over time.
We overlay the 3D representation of multiple timesteps, so a geometrically consistent scene appears as a sharp static background with the moving object tracing its trajectory through it.
MoGe4D generates almost no motion. 
In 4DNeX, background drift produces duplicated and smeared structure, while discontinuous object motion contributes to its high dynamic degree. 
Our overlays keep the background aligned across all timesteps while the dynamic content moves through it.

\subsection{Ablation Studies}
\label{sec:exp_ablation}

\paragraph{Reward corruption test.}
FOCAL decreases monotonically as geometric corruption increases, so it can serve as a reliable signal for geometric consistency. 
Following SGC~\citep{dou2026measuring3dspatialgeometric}, we reconstruct real and generated videos, perturb the reconstruction, and evaluate our reward on it. 
We use four perturbations of the kind a valid camera motion cannot explain. 
\begin{wrapfigure}{r}{0.4\textwidth}
  \begin{center}
\vspace{-0.3cm}
    \includegraphics[width=0.4\textwidth]{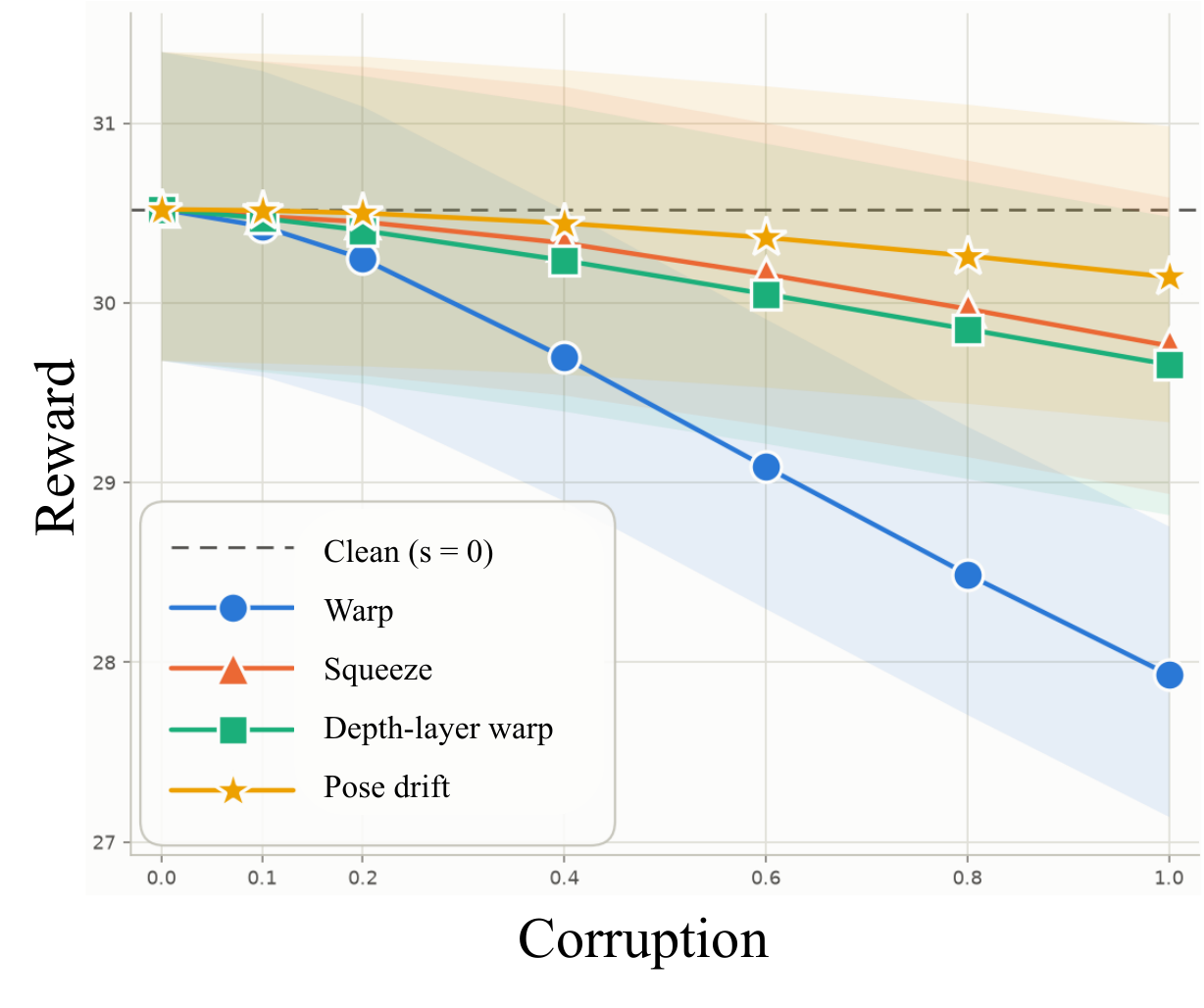}
  \end{center}
 \vspace{-0.6cm}
  \caption{Analysis of reward sensitivity to geometric corruption.}
  \vspace{-0.4cm}
  \label{fig:corruption_test}
\end{wrapfigure} 
Warp shifts the depth by a sinusoidal displacement field, deforming the scene non-rigidly. 
Squeeze stretches the depth map vertically while leaving the intrinsics unchanged. 
Depth-layer warp shifts the depth horizontally, which inverts parallax. 
Pose drift rotates the camera poses while leaving the depth untouched. 
The static-region mask is transformed together with the depth, so that changes in mask quality do not affect the measurement. 
As shown in Figure~\ref{fig:corruption_test}, the reward decreases as all four corruptions increase, confirming that our proposed reward measures geometric inconsistency rather than image fidelity.
Lines show mean rewards, shaded regions indicate $95\%$ bootstrap confidence intervals of the mean.

\begin{wrapfigure}{r}{0.4\textwidth}
  \begin{center}
\vspace{-0.5cm}
    \includegraphics[width=0.4\textwidth]{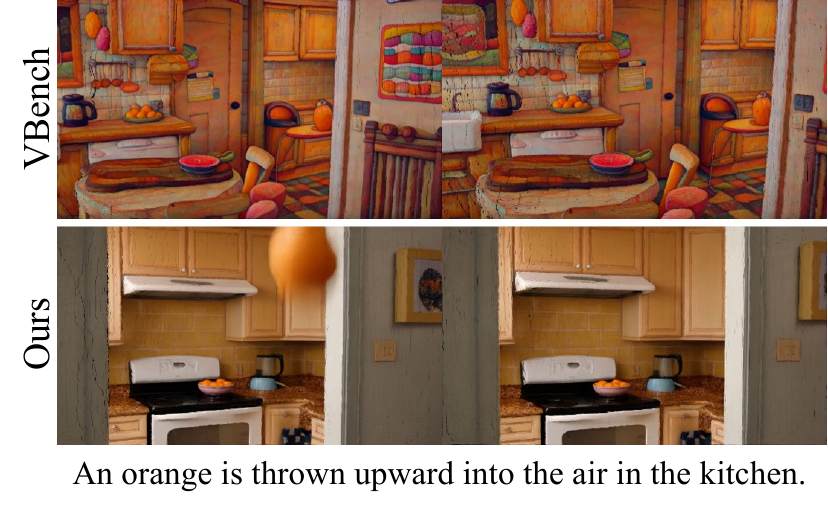}
  \end{center}
 \vspace{-0.6cm}
  \caption{VBench reward comparison.}
  \vspace{-0.4cm}
  \label{fig:abl_vbench}
\end{wrapfigure} 
\paragraph{Geometric reward vs.\ video-quality reward.}
\label{sec:abl_vbench}
VBench rewards capture general video quality but are not designed to assess geometric consistency, and directly optimizing them can lead to degenerate solutions. 
Following DDRL~\citep{ye2025dataregularizedreinforcementlearningdiffusion}, we train the VDM with direct reward fine-tuning on five VBench~\citep{huang2023vbenchcomprehensivebenchmarksuite} rewards. 
The VBench-trained model produces saturated videos (Figure~\ref{fig:abl_vbench}), and training longer adds blur and grid artifacts. 
The failure follows from what these rewards measure. 
Several VBench metrics reward high DINO or CLIP feature similarity between frames, collapsing inter-frame variation at the cost of blur and grid artifacts. 
Our geometric reward, in contrast, improves consistency and even video quality, which supports our claim that correcting geometric defects is an effective route to better video quality.
Additional details are in Appendix~\ref{sec:supp_vbench}.

\paragraph{Effect of FOV calibration.}
\label{sec:fov}
FOV calibration makes our reward invariant to camera motion, whereas other direct and indirect rewards correlate with it. 
Existing measures of geometric consistency are easier to satisfy when camera moves less, so these scores can be increased without improving geometry at all. 
As in Figure~\ref{fig:camera_corr}, direct rewards such as cross-view depth reprojection and input-view RGB reprojection~\citep{go2026textto3dstitchingmultiviewreconstruction,du2026videogpadistillinggeometrypriors}, as well as indirect rewards such as PickScore, all vary systematically with camera motion.

\begin{wraptable}{r}{0.3\columnwidth}
    \centering
    \scriptsize
    \vspace{-0.7cm}
    \caption{Effect of FOV calibration.}
    \label{tab:fov_calibration}
    \begin{tabular}{c|cc}
        \toprule
        $\gamma$ & VBench Avg. & Dynamic \\
        \midrule
        $0$ & 0.6802 & 0.395 \\
        \rowcolor{blue!10} $0.7$ (Ours) & \textbf{0.6861} & \textbf{0.425} \\
        $1$ & 0.6823 & \textbf{0.425} \\
        \bottomrule
    \end{tabular}
\end{wraptable}

FOV calibration prevents reward optimization from suppressing camera movement during training. Removing the calibration ($\gamma{=}0$) loses dynamic motion, since the model reduces camera motion to collect the easier reward (Table~\ref{tab:fov_calibration}). Our setting even matches the dynamic degree of a larger $\gamma$ ($\gamma{=}1$). Our calibrated setting achieves the highest video quality represented by VBench, indicating the best balance between geometric consistency and dynamic motion.

\section{Conclusion}
In this paper, we presented FOCAL, a geometric consistency reward for video diffusion models, and GeoFix, which applies FOCAL to reward alignment methods to improve geometric
consistency and visual quality.
Our reward evaluates cross-view depth consistency in static scene regions, targeting structural inconsistencies while preserving valid object motion. 
The core of our method is calibrating depth agreement against the expected value at a given camera overlap, reducing the extent to which greater overlap alone is rewarded as an improvement in geometric consistency.
FOCAL is computed directly from video latents through a stitched 4D reconstructor, without VAE decoding.
Experiments across training and inference-time reward alignment demonstrate improvements in geometric consistency, video quality and camera accuracy. 
We further show that GeoFix produces coherent 4D scenes compared to baselines.
Overall, GeoFix advances video diffusion models as priors for visually rich and geometrically coherent dynamic scene generation.


\bibliography{iclr2027_conference}
\bibliographystyle{iclr2027_conference}

\clearpage
\appendix
\section{Implementation Details}
\label{sec:supp_impl}

\subsection{Experimental Setup}

\subsubsection{Tasks and Datasets}
All three tasks produce a 4D Gaussian Splatting (4DGS) scene from a video diffusion model (VDM). In each case, a NeoVerse~\citep{yang2026neoverseenhancing4dworld} 4D reconstructor, stitched to the VDM latent space, decodes the VDM latents into a 4DGS scene.

\noindent\textbf{Text-to-4DGS (T2V-CompBench).}
We use 200 T2V-CompBench~\citep{sun2025t2vcompbenchcomprehensivebenchmarkcompositional} prompts and generate one video per prompt.
At generation time, each prompt is wrapped in a camera-motion sentence chosen at random from a fixed template set. This gives the 4D reconstructor parallax to recover geometry from. The augmentation is applied identically to every method.

\noindent\textbf{Image-to-4D generation (VBench-I2V).}
We use the 100 image--caption pairs of the VBench-I2V benchmark~\citep{huang2024vbenchcomprehensiveversatilebenchmark}. The input image is used as the first frame, and we apply camera prompt augmentation. For evaluation, we render every generated 4D scene along the same ReCamMaster novel-view camera trajectory.

\noindent\textbf{Camera-controlled image-to-4D generation.}
We use the 355 image--caption pairs of the VBench-I2V benchmark~\citep{huang2024vbenchcomprehensiveversatilebenchmark} and assign each image one of ten target camera trajectories (Table~\ref{tab:supp_traj}).
The pan trajectories translate the camera at a constant rate in the image plane. The zoom trajectories translate it along the optical axis at twice the pan rate.
Every method uses the same image--trajectory assignment.

\begin{table}[h]
\centering
\small
\caption{Ten target camera trajectories used for camera-controlled image-to-4DGS evaluation on 355 VBench-I2V images.}
\label{tab:supp_traj}
\begin{tabular}{ll}
\toprule
Trajectory & Trajectory \\
\midrule
Pan Left & Pan Left-Up \\
Pan Right & Pan Left-Down \\
Pan Up & Pan Right-Up \\
Pan Down & Pan Right-Down \\
Zoom In & Zoom Out \\
\bottomrule
\end{tabular}
\end{table}

\subsubsection{Baselines}
\noindent\textbf{Training-time alignment.}
We compare against the base Wan2.2-TI2V-5B~\citep{wan2025wanopenadvancedlargescale} without fine-tuning and three fine-tuned baselines: supervised fine-tuning (SFT) only, VIST3A~\citep{go2026textto3dstitchingmultiviewreconstruction} with SFT, and World-R1~\citep{wang2026worldr1reinforcing3dconstraints} with SFT.
For a controlled comparison, all fine-tuned models, including ours, use the same SFT dataset, and the same training GPU hours, and are decoded by the same frozen stitched 4D reconstructor. We use the same frozen stitched 4D reconstructor for reward computation in VIST3A and World-R1.

\noindent\textbf{Image-to-4D generation.}
We use the publicly released checkpoints of MoGe4D~\citep{zhang2026geometryawaresingleimage4dsynthesis} and 4DNeX~\citep{chen20254dnexfeedforward4dgenerative}. Both baselines generate their respective 4D point-cloud representations at 512$\times$512 resolution. Since their output representations differ from our 4DGS representation, we render all methods under a common target camera trajectory for evaluation. We derive the target camera trajectory from ReCamMaster. The predicted camera poses are first converted to the OpenCV camera convention and expressed as relative poses with respect to the first frame. The same relative camera motion is then applied consistently to all frames and methods. To account for the scale ambiguity of monocular 4D reconstruction, we normalize the translational magnitude of the camera trajectory such that it corresponds to 0.25 times the median scene depth. When projecting the 4D point clouds to the target views, multiple points may map to the same image pixel; in such cases, we use a standard depth test and retain the point closest to the camera.
 
The rendering procedure is adapted to the representation produced by each baseline. 4DNeX predicts a separate point map for each frame together with a corresponding input-view camera. For each frame, we transform its point map from the original input-view camera to the target ReCamMaster camera and rasterize the warped points in the target view. In contrast, MoGe4D represents the dynamic scene as a set of points initialized from the pixels of the first frame and tracked throughout the entire 41 frame sequence, rather than predicting an independent point map for every frame. As MoGe4D directly generates the dynamic point-cloud representation without an explicit per-frame input camera, we render the tracked 3D point positions at each frame directly from the corresponding target camera pose.

\noindent\textbf{Test-time steering.}
Test-time steering uses no trained checkpoint. We apply every steering method to the base camera-control model Wan2.2-Fun-5B-Control-Camera and compare unguided sampling (None), Best-of-$N$ (BoN) and DPS~\citep{chung2023diffusion}.

\subsubsection{Training}
We fine-tune Wan2.2-TI2V-5B with LoRA~\citep{hu2021loralowrankadaptationlarge} (rank 4, $\alpha{=}8$) on the query, key, value, and output projections of both the self- and cross-attention layers. The stitched 4D reconstructor is kept frozen.
Each optimizer step combines a supervised loss with an online reinforcement-learning loss based on DiffusionNFT~\citep{zheng2025diffusionnft}.

\noindent\textbf{Supervised term.}
We use the MultiCamVideo~\citep{bai2025recammastercameracontrolledgenerativerendering} (120{,}460 multi-camera videos) and the short-clip split of UltraVideo~\citep{xue2025ultravideo} (42{,}184 clips). Each video is 13 frames at 480$\times$832. With probability 0.5, the first frame is used as an image condition, so the model is trained for both text-to-video and image-to-video generation.

\noindent\textbf{Reinforcement-learning term.}
Each step draws 8 prompts from the VideoUFO~\citep{wang2025videoufomillionscaleuserfocuseddataset} training prompts and samples a group of 8 rollouts per prompt (64 rollouts per step). Each rollout has 13 frames and is sampled with 10 denoising steps and no classifier-free guidance. Each rollout is scored by our reward with $\gamma{=}0.7$.
The KL coefficient is $\beta_{\text{KL}}{=}5{\times}10^{-3}$.

\noindent\textbf{Optimization.}
We use AdamW~\citep{loshchilov2019decoupledweightdecayregularization} with a learning rate of $1{\times}10^{-4}$, $(\beta_1,\beta_2){=}(0.9,0.999)$, a weight decay of $1{\times}10^{-4}$, gradient clipping at 1.0, and bf16 mixed precision, and we train for 1,000 steps.

\subsubsection{Inference}
Unless stated otherwise, all videos are generated at a resolution of 480$\times$832 with 41 frames, 50 sampling steps, a classifier-free guidance scale of 5.0, and a fixed random seed shared across methods. For image-to-4DGS generation, the input image is fixed to the first latent frame at every denoising step. The resulting VDM latents are decoded into a 4DGS representation by the stitched reconstructor at a resolution of 518$\times$294.

\subsubsection{Test-Time Steering}
All steering methods share one fully stochastic flow-DDIM sampler (50 steps, 41 frames). The initial latent of each sample is derived deterministically from the run seed and the sample index. As a result, every method starts from the same noise and draws the same transition noise.
All methods maximize the proposed reward FOCAL, with $\gamma{=}0.7$. We use the following settings:
\begin{itemize}
    \item \textbf{BoN}: sample $N{=}20$ full trajectories and keep the highest-reward one.
    \item \textbf{DPS}~\citep{chung2023diffusion}: at each guided step, we back-propagate the reward through the transformer's clean-latent prediction $\hat{z}_{0|t}$, the stitched 4D reconstructor, and the depth-reprojection and FOV-overlap terms, and update the latent. We guide denoising steps 3--17 of 50 (15 guided steps). The static-region mask and the normalizer $D_{\max}$ are treated as constants. Since the hard FOV overlap $\Omega$ has zero gradient almost everywhere, we keep the exact $\Omega$ in the forward pass so that the reward value is unchanged, while gradients flow through a soft relaxation $\tilde{\Omega}$ into the predicted camera poses and intrinsics (see below).
\end{itemize}

\paragraph{Differentiable FOV overlap for first-order reward alignment methods.}
For a sphere sample $\mathbf{s}$ (Eq.~\ref{eq:supp_sphere}) expressed in camera $f$'s coordinates,
with horizontal angle $\mathrm{yaw}_f(\mathbf{s})$ and vertical angle $\mathrm{pitch}_f(\mathbf{s})$, we replace the
hard membership $\mathbf{s}\in\mathcal{A}_f$ with
\begin{equation}
a_f(\mathbf{s}) = \sigma\!\left(\frac{\theta^h_f/2 - |\mathrm{yaw}_f(\mathbf{s})|}{\tau}\right)
         \sigma\!\left(\frac{\theta^v_f/2 - |\mathrm{pitch}_f(\mathbf{s})|}{\tau}\right),
\end{equation}
where $\theta^h_f, \theta^v_f$ are the FOVs of Eq.~\ref{eq:supp_fov_angle}, $\sigma$ is the sigmoid and
$\tau = 0.5^\circ$. The intersection and union in Eq.~\ref{eq:supp_area_iou} become
\begin{equation}
\tilde{\Omega} = \frac{\sum_{\mathbf{s}} w(\mathbf{s})\, a_1(\mathbf{s})\, a_F(\mathbf{s})}
                      {\sum_{\mathbf{s}} w(\mathbf{s})\,\big(a_1(\mathbf{s}) + a_F(\mathbf{s}) - a_1(\mathbf{s})\, a_F(\mathbf{s})\big)},
\end{equation}
where the sums run over all sphere samples and $w(\mathbf{s}) = \cos e$ is the area weight of Eq.~\ref{eq:supp_area_iou}, with $e$ the elevation of $\mathbf{s}$ (Appendix~\ref{sec:supp_fov_overlap}).
We use a straight-through estimator: the forward pass keeps the exact $\Omega$, so the reward value
is identical, while gradients flow through $\tilde{\Omega}$ into the
predicted camera poses and intrinsics. The sphere radius $r$ is computed from detached depths.

\subsubsection{Stitched Model Training}

The stitched model feeds the latents of a frozen Wan~2.2~\citep{wan2025wanopenadvancedlargescale} video VAE encoder into NeoVerse~\citep{yang2026neoverseenhancing4dworld}. We initialize the stitching layer following VIST3A~\citep{go2026textto3dstitchingmultiviewreconstruction}, which gives the lowest regression error among the candidate encoder blocks. We train the stitched model by output distillation from the frozen NeoVerse. The student matches all of the teacher's outputs with $\ell_1$ losses: point maps, camera poses and intrinsics, depth, Gaussian parameters, and, for dynamic scenes, velocities. We train the stitching layer and the prediction heads fully, and adapt the NeoVerse backbone with LoRA (rank 64).

We first train on static scenes from DL3DV~\citep{ling2024dl3dv}. We then fine-tune on dynamic videos from UltraVideo~\citep{xue2025ultravideo}. The number of input frames per iteration is randomly sampled from lengths of the form $4n{+}1$. We use AdamW with a learning rate of $10^{-4}$.

\subsection{Evaluation Details}

\subsubsection{Evaluated Video}
Except for the VGGRPO comparison (Section~\ref{sec:supp_vggrpo}), we score rendered videos rather than raw VDM decoded outputs. For text-to-4DGS and test-time steering, the generated 4DGS scene is rendered using the input-view cameras predicted by the 4D reconstructor. For image-to-4D generation, we render every generated 4D scene along the same novel-view camera trajectory. Following the baseline protocol, we fill disoccluded regions with MoGe4D's 4D-ViSM, applied identically to every method. Scoring the rendered videos exposes geometry that is inconsistent across frames.

\subsubsection{Comparison Protocol with VGGRPO}
\label{sec:supp_vggrpo}
VGGRPO~\citep{an2026vggrpoworldconsistentvideogeneration} has released neither code nor checkpoints, so we compare with the numbers reported in its paper, matching its prompt and generation protocol on the VAE-decoded videos. Both methods build on Wan2.2 as the base model, and we apply the same setting identically to the base model and to ours. Because VGGRPO was evaluated with a different random seed, we compare the relative improvement of each method over its own base model rather than the absolute scores.

\subsubsection{Metrics per Task}
\begin{itemize}
    \item \textbf{Text-to-4DGS}: FOCAL and WorldScore.
    \item \textbf{Image-to-4D generation}: VBench subject consistency, temporal flickering, dynamic degree, and aesthetic quality; CLIP score for text alignment; and VideoPhy-2 physical commonsense (PC) for physical plausibility.
    \item \textbf{Test-time steering}: camera accuracy and WorldScore.
\end{itemize}

\subsubsection{Metric Definitions}
We compute metrics with the official implementations\footnote{\url{https://github.com/Vchitect/VBench}}\footnote{\url{https://github.com/haoyi-duan/WorldScore}}\footnote{\url{https://github.com/Hritikbansal/videophy}}:

\begin{enumerate}
    \item \textbf{VBench}~\citep{huang2023vbenchcomprehensivebenchmarksuite}: We use VBench in custom-input mode and report eight dimensions: subject consistency (DINO~\citep{caron2021emerging} feature similarity across frames), background consistency (CLIP feature similarity across frames), aesthetic quality (LAION aesthetic predictor), imaging quality (MUSIQ), overall consistency (ViCLIP text--video similarity), temporal flickering, motion smoothness (AMT-based frame interpolation prior), and dynamic degree (RAFT optical flow). We report the official \emph{Quality Score}, the normalized weighted average of the seven quality dimensions. In the text-to-4DGS setting of Table~\ref{tab:t2v_quantitative_comparison}, GeoFix improves the Quality Score from 0.7272 (Base) to 0.7588 (+4.3\%). We report the unweighted mean of all eight dimensions (\emph{VBench Avg.}) in Table~\ref{tab:fov_calibration}. Table~\ref{tab:video_quality_metrics} reports individual dimensions following the VGGRPO protocol, and Section~\ref{sec:exp_4d} uses subject consistency, temporal flickering, dynamic degree, and aesthetic quality.
    \item \textbf{WorldScore}~\citep{duan2025worldscoreunifiedevaluationbenchmark}: We report eight aspects: content alignment (CLIP score), 3D consistency (DROID-SLAM reprojection error), photometric consistency (SEA-RAFT optical-flow AEPE), style consistency (Gram-matrix distance of VGG-19 features), subjective quality (CLIP-IQA+ and CLIP aesthetic score), motion accuracy, motion magnitude (optical-flow magnitude), and motion smoothness (VFIMamba frame-interpolation error measured with LPIPS and SSIM).
    Motion accuracy measures whether motion is confined to dynamic regions. Upstream WorldScore obtains the dynamic region with a text-grounded detector. Our prompts do not name a single dynamic object, so we instead use SegAnyMo~\citep{huang2025segmentmotionvideos} to segment moving content in each rendered video and score flow inside and outside the resulting mask. 
    We omit camera control and object control, which require per-instance camera and object annotations that our benchmarks do not provide.
\item \textbf{VideoPhy-2}~\citep{bansal2025videophy2challengingactioncentricphysical}: We use VideoPhy-2-AutoEval (\texttt{videophy\_2\_auto}), an mPLUG-Owl-Video model with a LLaMA-7B backbone, further fine-tuned from VideoCon-Physics on VideoPhy-2 human annotations. We evaluate 32 sampled frames with the official physical-commonsense template (Figure~\ref{fig:supp_videophy2}) using greedy decoding. We report the mean physical commonsense (PC) score on a 1--5 scale.
    \item \textbf{CLIP score}~\citep{radford2021learning}: The cosine similarity between CLIP ViT-L/14 image and text embeddings, scaled by CLIP's logit scale, computed on every frame and averaged over frames.
    \item \textbf{Camera accuracy}: We compare the per-frame cameras predicted by the 4D reconstructor with the target trajectory. Predicted poses are defined only up to a similarity transform, so we express both trajectories relative to their first frame and report three errors: rotation error (the mean geodesic distance between the relative rotations, in degrees), translation-direction error (the median angle between the relative translation vectors, in degrees), and normalized ATE (the RMSE of the camera centers after a least-squares scale fit, divided by the maximum extent of the target trajectory). We use the median for translation-direction error because the first frames have almost no translation, which makes their direction ill-conditioned.
\end{enumerate}

\begin{figure}[!ht]
  \centering
  \small
  \begin{tcolorbox}[
    colback=gray!5!white,
    colframe=black,
    title=\textbf{Prompt for VideoPhy-2 Evaluation}
  ]
\textbf{Physical commonsense (PC).}\\
The following is a conversation between a curious human and an AI assistant. The assistant gives helpful, detailed, and polite answers to the user's questions.\\
Human: \texttt{<|video|>}\\
Human: Does this video adhere to the physical laws? Rate the video on a scale from 1 to 5, where 5 means full compliance and 1 means significant violations.\\
AI:
  \end{tcolorbox}
  \caption{Official VideoPhy-2 physical-commonsense (PC) query template used with \texttt{videophy\_2\_auto}.}
  \label{fig:supp_videophy2}
\end{figure}

\section{Stitched Model Reconstruction Results}

\paragraph{Setup.}
We evaluate on a subset of MultiCamVideo~\citep{bai2025recammastercameracontrolledgenerativerendering} dataset, which provides 10 synchronized cameras per scene.
\emph{Novel-time synthesis}: given the 33 even-indexed frames of one
camera, the model renders the 32 held-out odd-indexed frames.
\emph{Novel-view synthesis}: given 21 frames from a context camera, the
model renders the synchronized frames of a different, randomly chosen
target camera. We compare against 4DGT~\citep{xu20254dgtlearning4dgaussian}, MoVieS~\citep{lin2026moviesmotionaware4ddynamic}, and NeoVerse~\citep{yang2026neoverseenhancing4dworld}. NeoVerse and our model do not take ground-truth cameras as input, and
they predict poses in their own coordinate frame. For both models we
therefore follow the evaluation-time pose alignment of
NoPoSplat~\citep{ye2024poseproblemsurprisinglysimple}.

\paragraph{Results.}
As shown in Table~\ref{tab:novel_time_view}, our stitched model
outperforms baselines in most of the metrics in both novel-time and novel-view synthesis evaluation.
On novel-time synthesis, compared with NeoVerse, our model improves PSNR by $+1.68$\,dB, SSIM by $+0.066$, and LPIPS by $-0.031$. Since both models share the same backbone, we
attribute this gain to the spatio-temporally compressed video VAE
latents, which yield a more temporally coherent reconstruction.
Novel-view synthesis results show that operating on video latents does not degrade cross-view geometry.

\begin{table}[t]
    \centering
    \caption{Quantitative comparison on novel-time and novel-view reconstruction.}
    \label{tab:novel_time_view}
    \resizebox{\textwidth}{!}{
    \begin{tabular}{lccc|ccc}
        \hline
        Method & \multicolumn{3}{c|}{Novel Time} & \multicolumn{3}{c}{Novel View} \\
        & PSNR & SSIM & LPIPS $\downarrow$ & PSNR & SSIM & LPIPS $\downarrow$ \\
        \hline
        4DGT & 17.72 & 0.5510 & 0.5032 & 16.35 & 0.5015 & 0.5167 \\
        MoVieS & 19.12 & 0.5819 & 0.3785 & 16.83 & 0.4904 & 0.4271 \\
        NeoVerse & 20.30 & 0.5793 & 0.2917 & 17.23 & 0.5117 & \textbf{0.3716} \\
        Ours & \textbf{21.98} & \textbf{0.6451} & \textbf{0.2605} & \textbf{17.65} & \textbf{0.5187} & 0.3748 \\
        \hline
    \end{tabular}
    }
\end{table}

\section{Additional Results}

\subsection{Additional 4DGS Qualitative Results}

\begin{figure}[h]
    \centering
    \includegraphics[width=\linewidth]{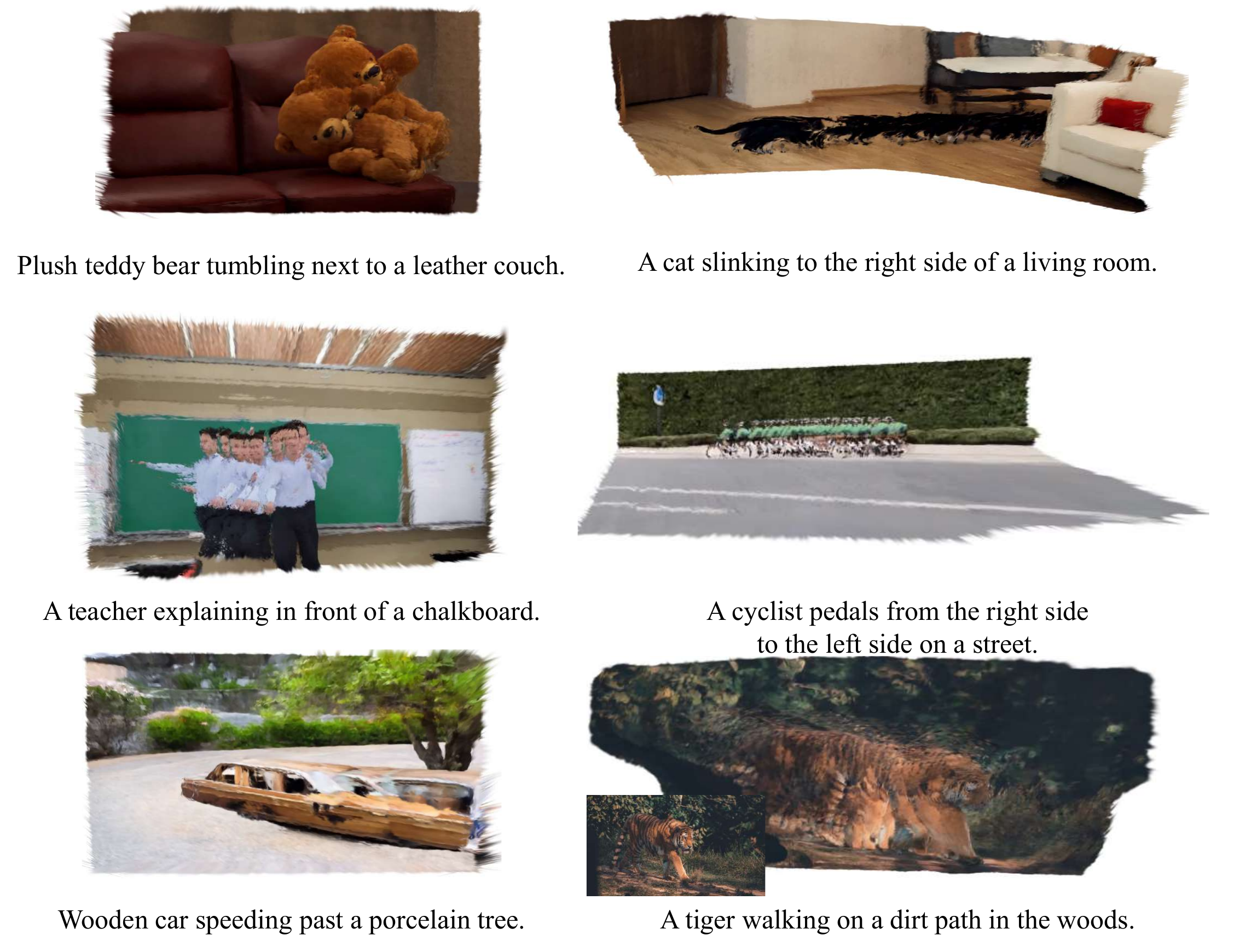}
    \vspace{-0.7cm}
    \caption{Additional qualitative results on Text-to-4DGS generation. We overlay 3D representations from multiple timesteps in a single view to visualize dynamic motion over time.}
    \label{fig:supple_qualitative_4dgs}
\end{figure}

Figure~\ref{fig:supple_qualitative_4dgs} shows additional Text-to-4DGS results.

\subsection{Additional Text-to-Video Qualitative Results}

\begin{figure}[h]
    \centering
    \includegraphics[width=\linewidth]{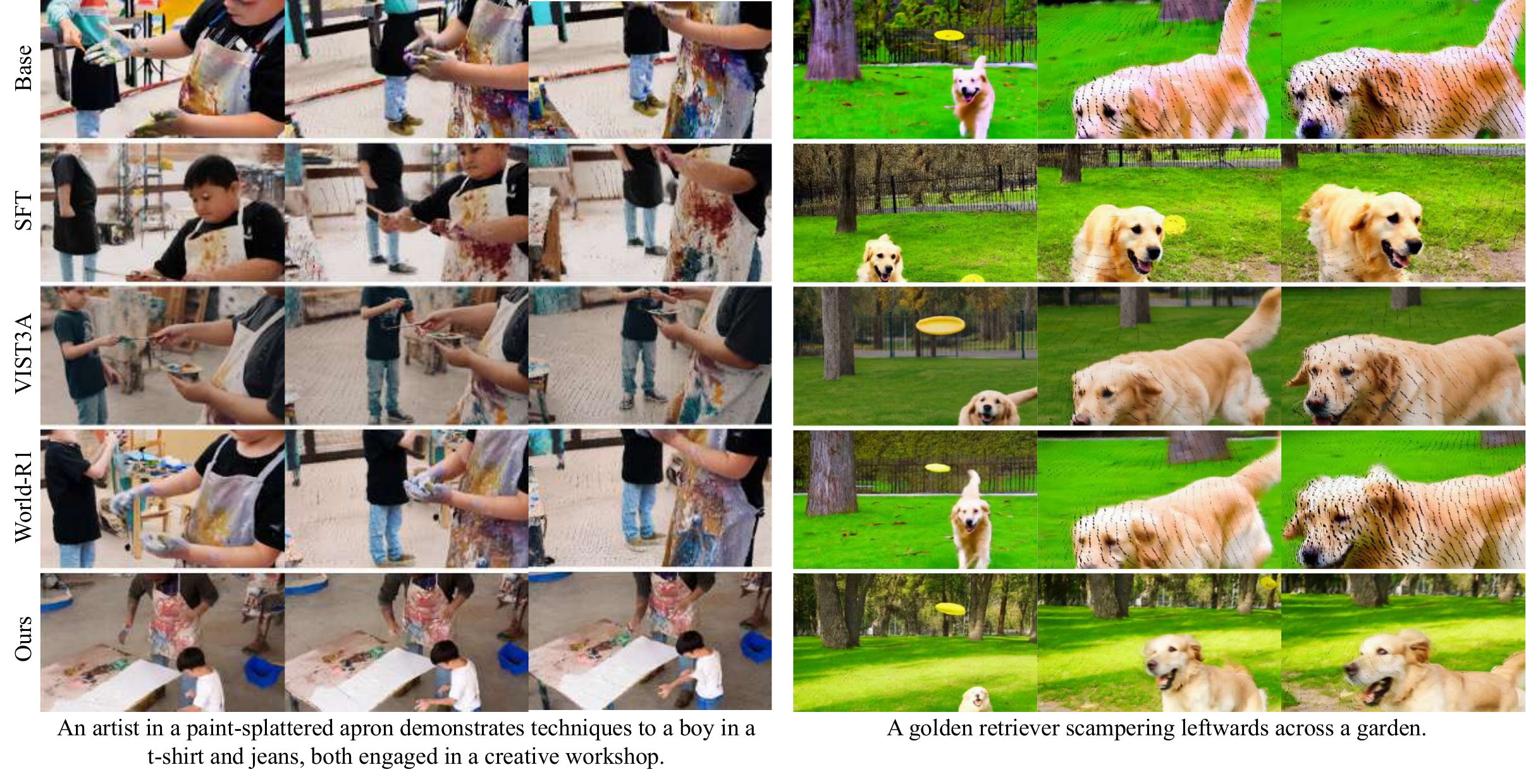}
    \vspace{-0.7cm}
    \caption{Additional qualitative comparison on Text-to-Video generation.}
    \label{fig:supple_qualitative_t2v}
\end{figure}

Figure~\ref{fig:supple_qualitative_t2v} provides additional 4DGS-rendered Text-to-Video comparisons against baseline methods.

\section{Additional Analysis}

\subsection{Overlap Bias in Depth Errors}
\label{sec:fov_analysis}
\begin{figure}[h]
    \centering
    \includegraphics[width=\linewidth]{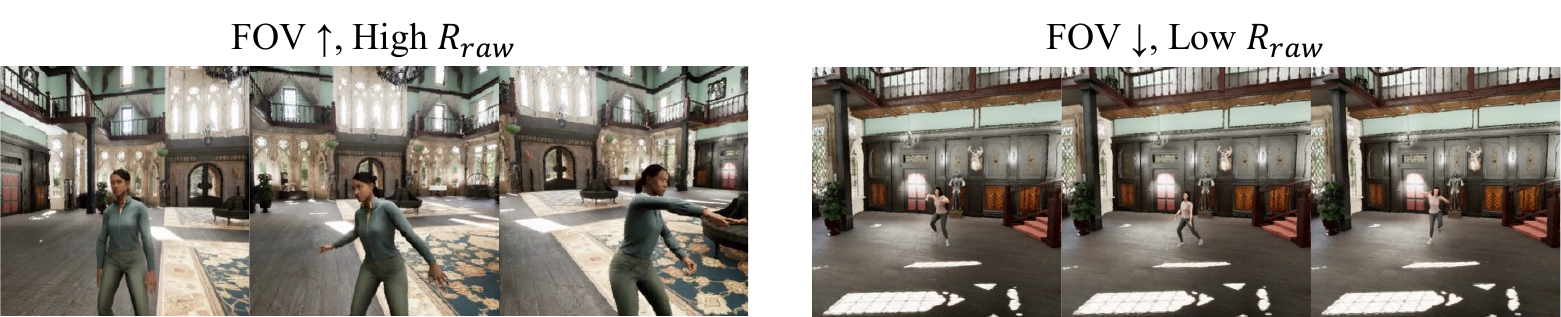}
    \vspace{-0.7cm}
    \caption{\textbf{Overlap bias in cross-view depth errors on reference videos.}
    Reference videos with a large FOV change (low overlap $\Omega$) exhibit larger depth reprojection errors than those with a small FOV change (high $\Omega$), even when their reconstruction quality is comparable.
    Thus, the raw depth-agreement score is influenced not only by geometric quality but also by camera motion.}    
    \label{fig:observation}
\end{figure}

Figure~\ref{fig:observation} visualizes the overlap bias discussed in Section~\ref{sec:observation} on the MultiCamVideo~\citep{bai2025recammastercameracontrolledgenerativerendering} dataset.
For each clip, we compute the raw cross-view depth reprojection error (Eq.~\ref{eq:reproj_err}) with the stitched reconstructor and the FOV overlap $\Omega$ (Eq.~\ref{eq:fov_overlap}).
Clips with large camera motion, and thus a large change in viewing frustum between the first and last frames (low $\Omega$), show substantially higher depth errors.
Clips with small camera motion (high $\Omega$) show low errors.
This gap arises from the viewing configuration rather than geometric inconsistency: with small overlap, reprojected points land near image boundaries, at grazing angles, or on newly revealed and occluded surfaces, where depth prediction is inherently harder.
A reward based on the raw score would therefore favor videos with little camera motion over videos that are more geometrically consistent, which motivates the calibration in Section~\ref{sec:3d_reward}.

\subsection{FOV Overlap Computation}
\label{sec:supp_fov_overlap}

\begin{figure}[h]
    \centering
    \includegraphics[width=\linewidth]{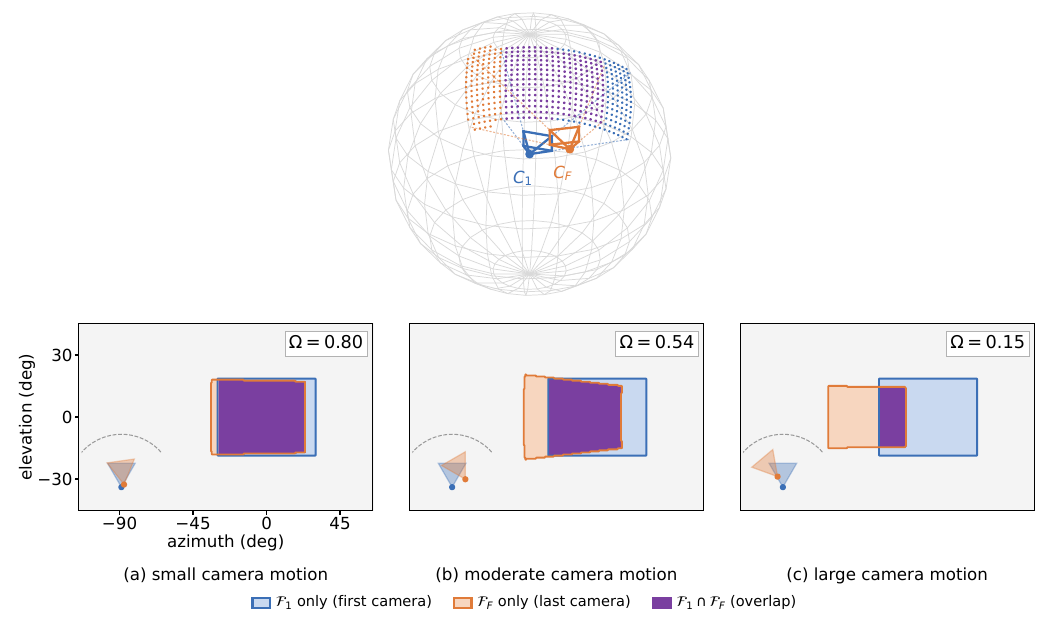}
    \vspace{-0.5cm}
    \caption{\textbf{FOV overlap computation. Top:} the first camera $C_1$ and the last camera $C_F$ are expressed in the first camera's frame, and a reference sphere of radius $r$ is centered at $C_1$.
    Sphere samples inside the FOV of $C_1$ only, of $C_F$ only, and of both cameras are shown in blue, orange, and purple, respectively.
    \textbf{Bottom:} the same samples on the azimuth--elevation grid for (a) small, (b) moderate, and (c) large camera motion between the first and last frames, with the resulting overlap $\Omega$.
    The top row corresponds to (b).
    Insets show the camera poses from above.}
    \label{fig:fov_visualization}
\end{figure}

We describe how the FOV overlap $\Omega$ in Eq.~\ref{eq:fov_overlap} is computed from the stitched reconstructor's camera predictions.
Figure~\ref{fig:fov_visualization} illustrates the procedure.

\paragraph{Camera FOV.}
For frame $f$ with predicted intrinsics $\mathbf{K}_f$ and image resolution $W\times H$, the horizontal and vertical FOVs are
\begin{equation}
    \theta^{h}_f = 2\arctan\!\left(\frac{W}{2f_x}\right),
    \qquad
    \theta^{v}_f = 2\arctan\!\left(\frac{H}{2f_y}\right),
    \label{eq:supp_fov_angle}
\end{equation}
where $f_x$ and $f_y$ are the focal lengths in $\mathbf{K}_f$.
We use only the first and last frames for efficiency, $f\in\{1,F\}$, which span the full camera trajectory of the clip.

\paragraph{Reference sphere.}
We express both cameras in the coordinate frame of the first camera, $\tilde{\mathbf{T}}_f=\mathbf{T}_1^{-1}\mathbf{T}_f$, so that $C_1$ lies at the origin.
We then place a reference sphere of radius $r$ centered at $C_1$ (Figure~\ref{fig:fov_visualization}, top).
We set $r$ to the 75th percentile of the predicted depths of the clip, so that the sphere lies at a typical scene distance and camera translation is measured relative to the scene scale.
We use a percentile instead of the maximum or the mean because depth distributions are often heavy-tailed: regions such as the sky or distant backgrounds receive very large, and often unreliable, depth values.
A radius driven by these regions would place the sphere far beyond the actual scene content, so that even a large camera translation becomes negligible relative to $r$ and $\Omega$ reflects almost only rotation.
We sample the sphere on a uniform azimuth--elevation grid with $360\times180$ points,
\begin{equation}
    \mathbf{s}_{a,e}
    =
    r\,\bigl(\cos e\cos a,\;\cos e\sin a,\;\sin e\bigr),
    \qquad
    a\in[-\pi,\pi],\; e\in[-\tfrac{\pi}{2},\tfrac{\pi}{2}].
    \label{eq:supp_sphere}
\end{equation}

\paragraph{Area-weighted IoU.}
A uniform grid in azimuth and elevation oversamples the sphere near the poles, since the area of a grid cell is proportional to $\cos e$.
We therefore weight each point by $w(\mathbf{s}_{a,e})=\cos e$ and compute
\begin{equation}
    \Omega
    =
    \frac{\sum_{\mathbf{s}\in\mathcal{A}_1\cap\mathcal{A}_F} w(\mathbf{s})}
         {\sum_{\mathbf{s}\in\mathcal{A}_1\cup\mathcal{A}_F} w(\mathbf{s})},
    \label{eq:supp_area_iou}
\end{equation}
so that $\Omega$ approximates the ratio of the spherical areas of the intersection and union of the two FOVs.
$\Omega=1$ when the first and last cameras see the same region, and $\Omega=0$ when their FOVs do not overlap on the sphere.
As shown in Figure~\ref{fig:fov_visualization}(a)--(c), $\Omega$ decreases as the last camera rotates and translates further away from the first camera.

\subsection{Analysis of VBench Rewards}
\label{sec:supp_vbench}

\begin{figure}[h]
    \centering
    \includegraphics[width=\linewidth]{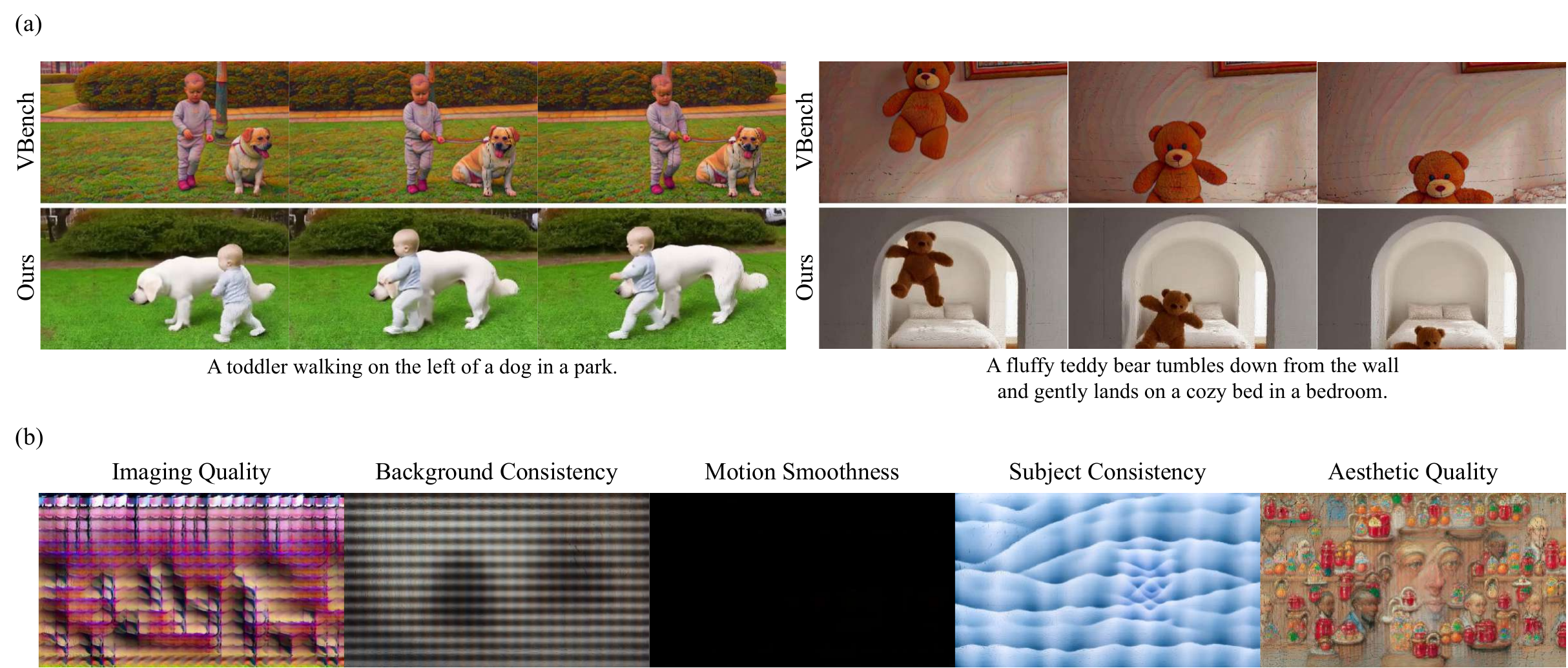}
    \vspace{-0.7cm}
    \caption{\textbf{Analysis of VBench rewards. (a)} Additional qualitative results for the video-quality reward ablation in Section~\ref{sec:exp_ablation}, comparing training with VBench rewards against FOCAL.
    \textbf{(b)} Representative artifacts from optimizing each of the five VBench rewards individually.}
    \label{fig:supple_vbench}
\end{figure}

We provide additional analysis of the video-quality reward ablation in Section~\ref{sec:abl_vbench}.
Figure~\ref{fig:supple_vbench}(a) shows additional qualitative results from training with the five VBench~\citep{huang2023vbenchcomprehensivebenchmarksuite} rewards used by DDRL~\citep{ye2025dataregularizedreinforcementlearningdiffusion}.
The VBench-trained model exhibits excessive saturation and unnatural color casts, whereas our geometric reward produces more natural colors and cleaner scene structure.
To examine the contribution of each reward, Figure~\ref{fig:supple_vbench}(b) shows results obtained by optimizing each of the five metrics individually in the direct reward fine-tuning setting.
The resulting artifacts reveal distinct failure modes, which we interpret below in terms of what each metric measures.

\paragraph{Imaging quality.}
Optimizing imaging quality produces strong color banding and repetitive, high-contrast textures that obscure the scene.
This reward averages the frame-wise predictions of MUSIQ~\citep{ke2021musiqmultiscaleimagequality}, an image-quality predictor trained to assess distortions such as blur, noise, and exposure errors.
The observed patterns suggest that direct optimization can exploit the predictor's learned responses to contrast and texture, amplifying artificial detail without improving perceptual quality.
A high predicted image-quality score does not impose explicit constraints on scene geometry or temporal coherence, allowing such structured artifacts to emerge.

\paragraph{Background consistency.}
Optimizing background consistency produces blurred content overlaid with a regular grid.
Despite its name, this metric compares global CLIP~\citep{radford2021learning} image features extracted from whole frames, without explicitly isolating the background.
It averages the CLIP feature similarity of each frame to the first frame and to its immediately preceding frame.
Consequently, reducing frame-to-frame feature variation can be favored even at the expense of image fidelity: blur can suppress changing details, while a persistent grid can provide a common appearance across frames.
This offers a shortcut to feature consistency without preserving a sharp, geometrically consistent background under camera motion.

\paragraph{Motion smoothness.}
Optimizing motion smoothness leads to an almost black, nearly featureless output.
The metric interpolates intermediate frames from neighboring frames and rewards a small pixel-wise difference between the interpolated and actual frames.
Temporally uniform, low-contrast content makes this prediction task easier: suppressing visible structures and their motion can reduce interpolation error without generating realistic dynamics.
The observed collapse is consistent with this incentive, since the reward itself does not require nontrivial motion or recognizable scene content.

\paragraph{Subject consistency.}
Optimizing subject consistency replaces recognizable objects with repetitive, smooth blue patterns.
This metric measures DINO~\citep{caron2021emerging} feature similarity to the first and preceding frames; it does not explicitly verify the presence or geometry of the intended subject.
Repeating similar structures across frames can therefore stabilize the descriptor even when subject identity is lost.
The resulting pattern illustrates how consistency in feature space can be achieved by suppressing meaningful appearance variation, rather than maintaining a coherent object as its pose or viewpoint changes.

\paragraph{Aesthetic quality.}
Optimizing aesthetic quality yields a collage-like image with repeated faces and decorative fragments instead of a coherent scene.
The reward averages a LAION~\citep{schuhmann2022laion5bopenlargescaledataset} aesthetic predictor applied to CLIP image embeddings of individual frames.
Such optimization can amplify visual motifs associated with high predicted aesthetic scores while leaving their spatial arrangement and temporal evolution unconstrained.

These examples illustrate the gap between evaluating generated videos with perceptual proxies and directly optimizing those proxies.
Temporal feature similarity can admit degenerate content, while frame-wise quality predictors can favor artifacts outside their intended image distribution.
Our reward instead evaluates cross-view geometric agreement in static regions, providing an explicit constraint on scene structure.
The comparison in Figure~\ref{fig:supple_vbench}(a) supports the finding in Section~\ref{sec:exp_ablation} that improving geometric consistency can also improve visual quality.

\section{Limitations}

Our method inherits the bias of the pretrained VDM and the 4D reconstruction model. The depth and camera predictions are distilled from a pretrained 4D reconstructor, so failure cases of the teacher can produce inaccurate reward signals.
Errors in the dynamic segmentation stage can penalize valid motion or leave inconsistent regions unconstrained. We expect that 4D reconstructors and dynamic tracking will further improve. 
Finally, computing the reward requires running the stitched reconstructor and the motion segmentation network, and gradient-based guidance additionally requires backpropagation through the reconstructor, which increases training and inference cost.

\end{document}